 \documentclass[serif,twocolumn,alpha-refs]{wiley-article}

\usepackage{siunitx}

\usepackage{graphicx} 
\usepackage{multirow}
\usepackage{multicol}
\usepackage{algorithm}
\usepackage{algpseudocode}
\usepackage{booktabs} 
\papertype{Original Article}
\paperfield{Journal Section}

\title{Fast and Incremental Loop Closure Detection with Deep Features and Proximity Graphs}

\author[1]{Shan An}
\author[1]{Haogang Zhu}
\author[2]{Dong Wei}
\author[3]{Konstantinos~A.~Tsintotas}
\author[3]{Antonios~Gasteratos}

\affil[1]{School of Computer Science and Engineering, Beihang University, Beijing, 100191, China.}
\affil[2]{Tech \& Data Center, JD.COM Inc., Beijing, 100108, China.}
\affil[3]{Department of Production and Management Engineering, Democritus University of Thrace, Xanthi 67132, Greece.}

\corraddress{Haogang Zhu, School of Computer Science and Engineering, Beihang University, Beijing, 100191, China.}
\corremail{haogangzhu@buaa.edu.cn}

\fundinginfo{The Major Program of the National Natural Science Foundation of China (91846201), Grant/Award Number: 91846201; The National Natural Science Foundation of China, Grant/Award Number: 72071069}

\runningauthor{An et al.}

\begin{document}

\begin{frontmatter}
\maketitle

\begin{abstract}
\color{black}
In recent years, the robotics community has extensively examined methods concerning the place recognition task within the scope of simultaneous localization and mapping applications.
\color{black}
This article proposes an appearance-based loop closure detection pipeline named ``FILD++" (Fast and Incremental Loop closure Detection).
First, the system is fed by consecutive images and, via passing them twice through a single convolutional neural network, global and local deep features are extracted.
Subsequently, a hierarchical navigable small-world graph incrementally constructs a visual database representing the robot's traversed path based on the computed global features.
Finally, a query image, grabbed each time step, is set to retrieve similar locations on the traversed route.
An image-to-image pairing follows, which exploits local features to evaluate the spatial information. \color{black}
Thus, in the proposed article, we propose a single network for global and local feature extraction in contrast to our previous work (FILD), while an exhaustive search for the verification process is adopted over the generated deep local features avoiding the utilization of hash codes.
\color{black} 
Exhaustive experiments on eleven publicly available datasets exhibit the system's high performance (achieving the highest recall score on eight of them) and low execution times (22.05 $ms$ on average in New College, which is the largest one containing 52480 images) compared to other state-of-the-art approaches.

\end{abstract}
\end{frontmatter}

\section{Introduction}
\label{sec:introduction}

Autonomous robots have to explore unknown areas while retaining the capability to construct a reliable map of the environment \citep{garcia2015vision, kostavelis2015semantic}.
This process is widely known as Simultaneous Localization and Mapping (SLAM) and constitutes an essential component for any modern robotic system~\citep{cadena2016past}.

Besides, place recognition --the ability to match a scene with a different one located about the same spot-- is necessary to generate a valid map~\citep{lowry2016visual}.
In recent years, the mobile robot platforms' increased computational power allowed cameras to be established as the primary sensor to perceive the appearance of a scene \citep{cummins2008fab, cummins2011appearance3, engel2015large, tsintotasRAL}.
However, the noisy sensor measurements, modeling inaccuracies, and errors due to field abnormalities affect the performance of SLAM.
Identifying known locations in the traversed route based on camera information to rectify the incremental pose drift is widely known as visual loop closure detection \citep{mei2010closing, zhang2011borf, botterill2011bag, tsintotas2018doseqslam, han2021novel}.
This operation is highly related to image retrieval, as the system tries to find the most similar visual entry within a visual database, which is explicitly built using camera measurements gathered along a trajectory.
There are two main stages in this process, namely \textit{filtering} and \textit{re-ranking} \citep{teichmann2019detect-to-retrieve}.
Regarding \textit{filtering}, the database elements are sorted according to their similarity to the query image, \textit{i.e.,} the current robot's view.
Then, during \textit{re-ranking}, each candidate image-pair generated from the \textit{filtering} is verified based on its spatial correspondences \citep{radenovic2018revisiting}.

Early studies in image retrieval used global description vectors, such as color or texture, to represent the visual data \citep{oliva2001modeling, torralba2003context, konstantinidis2005image, oliva2006building}.
The subsequent pipelines utilized the shape and local information extracted through point-of-interest detection and description methods to find the most similar candidates \citep{lowe2004distinctive, bay2006surf, calonder2010brief, amanatiadis2011evaluation, rublee2011orb}.
These approaches provided robust detection against rotation and scale changes.
However, the increased time needed to extract and match local features constitutes a significant bottleneck, particularly in highly textured environments~\citep{tsintotas2019appearance}.
Therefore, researchers adopt more sophisticated solutions to overcome this drawback, such as quantizing the descriptor space, producing more compact representations, and faster indexing.

The so-called \mbox{Bag-of-Words (BoW)} model~\citep{sivic2003video}, usually constructed through $k$-means clustering \citep{macqueen1967some}, employs the widely utilized Term-Frequency Inverse-Document-Frequency (TF-IDF) technique to generate visual words histograms that represent the camera data.
In many BoW-based place recognition approaches, the proper image-pair is retrieved via histogram comparisons~\citep{galvez2012bags, mur2014fast, bampis2016encoding, bampis2018fast, tsintotas2018seqslam}.
Such methods exhibit high accuracy and low execution times, which are achieved due to the utilization of indexing techniques, \textit{e.g.,} the hierarchical \mbox{$k$}-means tree \citep{nicosevici2012automatic}, \mbox{$k$-d} tree \citep{liu2012indexing}, and $k$-NN graph~\citep{hajebi2014efficient}.
Nevertheless, their functionality is highly dependent on the training environment wherein the visual data are extracted and, in turn, on the produced vocabulary.
Some visual loop closure detection frameworks incorporate mechanisms to map the environment through an incrementally generated visual vocabulary to cope with such dependencies \citep{filliat2007visual, angeli2008fast, labbe2013appearance, khan2015ibuild, tsintotas2018assigning}.
However, due to their database construction, these pipelines mainly adopt voting techniques to indicate the most similar location within the traversed route.

Compared to hand-crafted, the features  extracted from specific layers of Convolutional Neural Networks (CNNs) show high discrimination power \citep{babenko2014neural, sunderhauf2015performance, hou2015convolutional, gordo2016deep, anshan2019}.
Thus, CNN-extracted elements became a popular choice for many image classification \citep{krizhevsky2012imagenet} and scene recognition \citep{zhou2014learning} applications.
Afterward, the proper image is selected through comparison techniques similar to BoW schemes.
However, the spatial information embedded in image frames, which is crucial for data association between image-pairs for SLAM, is missing in the location's global representation.
Hence, methods for extracting local features have been developed.
DEep Local Feature (DELF), one of the original methods proposed for local CNN-based feature extraction, selects key-points based on an attention mechanism~\citep{noh2017large}.
Subsequently, the description stage is achieved using dense, localized features.
Finally, Principal Component Analysis (PCA) whitening reduces the descriptor space and improves the retrieval accuracy \citep{jegou2012negative}.

In our previous work \citep{anshan2019}, CNN-based global features extracted by MobileNetV2~\citep{sandler2018mobilenetv2} were fed into a Hierarchical Navigable Small World (HNSW) graph~\citep{malkov2018efficient} to map the environment via an incrementally generated visual database.
In addition, the graph allowed for short indexing times when searching for Nearest Neighbors (NN).
Then, local features, extracted via Speeded Up Robust Features (SURF)~\citep{bay2006surf}, were converted to binary codes to achieve real-time geometrical verification between the chosen image-pair.
In this work, a similar scheme for mapping the robot's traversed path has been adopted.
Besides, we utilize two forward passes through a single network for global and local features extraction.
The main advantages offered by this strategy are: (i) the highly reduced execution time for feature extraction and (ii) a significant accuracy improvement due to CNN-based features' better representation.
Furthermore, in this article, we introduce a \textit{re-ranking} optimization, which is based on local features' low dimensional space (40 bins).
Due to this fact, an exhaustive search is employed, unlike our previous work where hash codes were employed \citep{cheng2014fast}, to improve the verification process.
Finally, the proposed framework is more compact, simpler, and much faster than FILD~\citep{anshan2019}.
The presented algorithm is evaluated experimentally against a total of eleven benchmark datasets.
As a final note, the source code\footnote{https://github.com/AnshanTJU/FILD} of our Fast and Incremental Loop closure Detection (FILD) pipeline, dubbed as ``FILD++," is made publicly available to facilitate future studies.

The remainder of the paper is organized as follows: In Section \ref{sec:related}, a literature review of the most prominent works on visual loop closure detection is given.
Section \ref{ImageRepresentation} describes our deep features, and Section \ref{HNSW} introduces our HNSW visual database.
In Section \ref{pipeline}, the proposed detection pipeline is detailed, while the experimental protocol and the outcoming comparative results follow in Sections \ref{Experimental Protocol} and \ref{results}, respectively.
Finally, Section \ref{conclusion} discusses the proposed approach, draws conclusions, and provides our plans.

\section{Related Work}
\label{sec:related}

This section presents a literature review regarding the approaches which tackle the task of appearance-based loop closure detection.
Depending on their visual feature extraction techniques, these pipelines are distinguished into two categories: hand-crafted features and CNN-based features.

\subsection{Approaches using Hand-crafted Features}

Since many researchers quantize the extracted features to generate a visual vocabulary and cope with the large amount of features, off-line and incremental approaches are presented according to the process they follow to construct their database.
Fast Appearance-Based MAPping (FAB-MAP) is considered to be the most popular off-line approach~\citep{cummins2008fab, cummins2011appearance3}.
It uses a pre-trained SURF dictionary and a Chow Liu tree to learn its words' co-visibility \citep{chow1968approximating}.
BoWSLAM allows robots to navigate in unknown environments by utilizing the BoW feature matching with FAST corner detector and image patch descriptor \citep{botterill2011bag}.
\cite{galvez2012bags} proposed a hierarchical BoW model, built with local binary features in addition to direct and inverse indexes.
Their method was improved by employing ORB features~\citep{rublee2011orb} to incorporate rotation and scale invariance properties~\citep{mur2014fast}.
Similarly, previously visited locations were detected inside a Parallel Tracking and Mapping (PTAM) framework \citep{klein2007parallel}.
\cite{bampis2016encoding, bampis2018fast} combined the visual words' occurrences of sequence segments, \textit{i.e.,} groups-of-images, to assist the matching process.
Recently, points and lines were combined based on information entropy to realize accurate loop closure detection \citep{han2021novel}.

While the approaches mentioned above relied on a static visual vocabulary adapted to the training environment, in the work of \cite{angeli2008fast} an incrementally constructed vocabulary was proposed.
Loops were identified via the matching probability of a Bayesian scheme.
In a similar manner, an agglomerative clustering algorithm was adopted for database generation~\citep{nicosevici2012automatic}.
The stability between visual elements' associations was attained using an incremental image-indexing process in conjunction with a tree-based feature-labeling method.
Real-Time Appearance-Based Mapping (RTAB-Map) proposed a memory management mechanism to limit the number of candidate locations~\citep{labbe2013appearance}.
An Incremental bag of BInary words for Appearance-based Loop closure Detection (IBuILD) was proposed by~\cite{khan2015ibuild}.
Visual words were generated via feature matching on consecutive images, while a likelihood function decided about the location pairing.
Hierarchical Topological Mapping (HTMap) proposed by \cite{garcia2017hierarchical} relied on a loop closure scheme based on the Pyramid Histogram of Oriented Gradients (PHOG)~\citep{bosch2007representing}.
Similar locations are highlighted due to binary local features' correspondences.
An incremental approach exerting binary descriptors and dynamic islands was proposed in the work of \cite{garcia2018ibow}, while \cite{tsintotas2018assigning} dynamically segmented the incoming image stream to formulate places represented by unique visual words.
A probabilistic voting scheme followed, aiming to indicate the proper place, while an image-to-image pairing was held based on the locations' spatial correspondences.
The same authors, proposed a mapping algorithm based on an incrementally generated visual vocabulary constructed through local features tracking \citep{tsintotasRAL}.
The authors improved their method through the addition of a temporal filter and a vocabulary management technique in \citep{tsintotas2021modest}.
The candidate locations were chosen through their probabilistic binomial score~\citep{gehrig2017visual}.
A modified growing self-organizing network was proposed by \cite{kazmi2019detecting} for learning the topological representation of global gist features~\citep{oliva2001modeling}.

\subsection{Approaches using Convolutional Neural Networks Features}

The impressive performance of CNNs, exhibited on a wide variety of tasks, has been the main reason for their becoming the principal solution to many visual place recognition systems.
Utilizing an end-to-end trainable and generalized VLAD layer~\citep{jegou2010aggregating}, NetVLAD was proposed for similar locations' identification \citep{arandjelovic2016netvlad}.
A Spatial Pyramid-Enhanced VLAD (SPE-VLAD) layer was proposed by \cite{yu2019spatial} to encode the feature extraction and improve the loss function.
PCANet \citep{chan2015pcanet} employed a cascaded deep network to extract unsupervised features
improving the loop closure detection pipeline \citep{xia2016loop}.
\cite{cascianelli2017robust} proposed a visual scene modeling technique that preserved the geometric and semantic structure and, at the same time, improved the appearance invariance.
A multi-scale pooling exertion allowed for condition- and viewpoint-invariant features to be generated \citep{chen2017deep}.
Omnidirectional CNN was proposed to mitigate the challenge of extreme camera pose variations~\citep{wang2018omnidirectional}.
In the work of \cite{chen2018learning}, the authors proposed an attention mechanism capable of being incorporated into an existing feed-forward network architecture to learn image representations for long-term place recognition.
A useful similarity measurement for detecting revisited locations in changing environments was proposed by \cite{xin2019real}.
The combination of a neural network inspired by the Drosophila olfactory neural circuit with an \mbox{1-\textit{d}} Continuous Attractor Neural Network resulted into a compact system exhibiting high performance~\citep{chancan2020hybrid}.
Such works commonly used CNNs to extract the global descriptor of a scene, while few of them applied CNNs to extract local information for appearance-based loop closure detection.

\section{Image Representation}
\label{ImageRepresentation}

\begin{figure*}
	\begin{center}
		\includegraphics[width=0.95\linewidth]{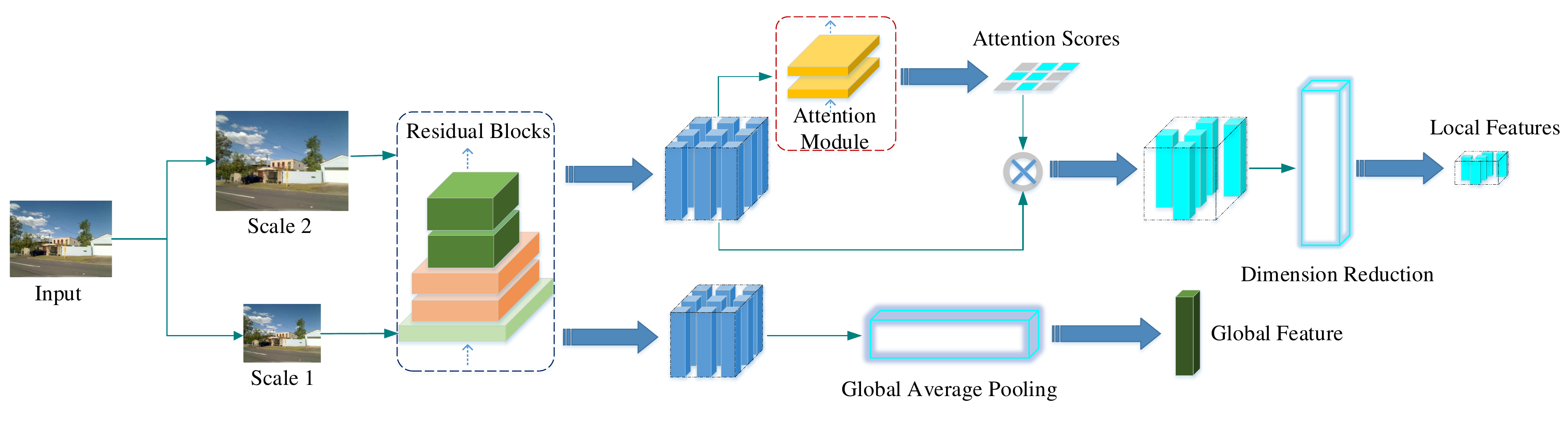}
	\end{center}
	\caption{The modified version of DEep Local Feature (DELF)~\citep{noh2017large} architecture for \textcolor{black}{feature extraction}.
		We extract the incoming visual stream's global and local representations via two passes of the proposed fully convolutional network.
		Three components constitute its structure, namely: the backbone, the global feature extraction branch and the local feature extraction branch.
		The first component is based on the residual blocks of ResNet50~\citep{he2016deep}.
		Simultaneously, for the global and local branches, an average pooling technique, an attention module and a dimension reduction method are used.
		The attention module is adopted for the corresponding scores' generation.
		This way, the most relevant features are assigned with higher scores prior to their dimensionality reduction.
		More specifically, the first-scale features are fed to the global branch, while the features from the second are sent to the local branch.
	}
	\label{fig:delf}
\end{figure*}

Our feature extraction module relies on a Fully Convolutional Network (FCN) to generate specific representations from the incoming image frames.
Aiming to achieve an enhanced image representation, the proposed architecture is implemented upon a modified version of DELF employing both global and local features in different scales through a double-pass process.
We choose the initial three convolutional blocks of ResNet50 \citep{he2016deep} as the backbone of our network, while the output of the last layer is fed into the feature extraction module as depicted in Fig. \ref{fig:delf}.
Aiming for compact and discriminative features, image-level labeled information is used to train the network.

\subsection{Incoming Image Frame}

To extract representative and robust features, we use different scales to extract the global and local features.
Contrary to the original version of DELF, where image pyramids are constructed using seven different scales, the proposed system employs only two of them; one for global and the other for local features' extraction.
According to the experiments, the scales for extracting global and local features are set as 0.5 and 1.4.
Our method performs even better when specific scales for different datasets are used.
However, the forward processing on multiple scales requires more times than a single scale.
Due to this fact, the selected parameters (see Section \ref{sec_scales}) provide a good trade-off between accuracy and timing, affording a system with low complexity and high performance.

\subsection{The Backbone Network}

Our backbone network comprises the first three convolutional blocks of ResNet50.
The initial block contains a \mbox{$7\times7$} convolution layer followed by a Batch Normalization (BN), a Rectified Linear Unit (ReLU), and a \mbox{$3\times3$} max-pooling layer with stride 2.
The second convolutional block includes three residual blocks each of which comprises three layers: \mbox{$1\times1$}, \mbox{$3\times3$}, and \mbox{$1\times1$}, respectively.
The \mbox{$1\times1$} convolution layers are used to reduce/increase the feature map's channels, while a BN and a ReLU follow each layer.
The final convolutional block comprises four residual blocks, which are similar to the previous block but are considered for the feature map's dual channels.

\subsection{Global Features}

A Global Average Pooling (GAP) layer~\citep{lin2013network} is applied to the output feature map $w\times{h}\times{c}$ of the backbone network to produce a single description vector for the incoming visual data.
Here, $w$, $h$, $c$ are feature map's width, height and channels, respectively.
As a result, the feature map's dimensionality is reduced to $1\times{1}\times{c}$ since GAP generates a single number per channel, which is the average of all $w\times{h}$ values.
GAP's output forms the employed global feature.

\subsection{Local Features}
\subsubsection{Attention-based local features}
Each pixel in the backbone network's output is considered as a local grid; the feature map is the dense sample of this grid.
The tensor composed of all grid channels is treated as a local feature, while the corresponding keypoint is located at the center of the receptive field in the pixel coordinates.

Since not all the densely extracted elements are appropriate for the intended recognition task, an attention module consisting of two $1\times{1}$ convolutional layers is applied to select a subset of them.
This module aims to learn a score function for each local feature and creates the corresponding score map of size $w\times{h}\times{1}$.
A softplus activation \citep{dugas2001incorporating} is deployed in the second layer to ensure the score is non-negative.
Then, the elements which present a value higher than a score threshold are selected.
It is noted that in this case, the local features are firstly computed and then selected.
This process differs from the hand-crafted techniques wherein the keypoints are firstly detected, and then their description vectors are generated.

\color{black}
The score map learning process is the same as the original version.
The features to be learned by the attention model are denoted as $f_n \in \mathbb{R}^d, n=1,...,N$, with $d$ is the feature dimension. 
The score function for each feature is $\alpha(f_n;\theta)$, with $\theta$ denoted the paramter of function $\alpha(\cdot)$.  
The network generates the output logit $\mathrm{y}$ by a wesighted sum of the feature vectors:
\begin{equation}
	\mathrm{y}=\mathrm{\textbf{W}}(\sum_{n}\alpha(f_n;\theta) \cdot f_n)
	\label{eq01}
\end{equation}
$\mathrm{\textbf{W}}\in \mathbb{R}^{M \times d}$ is the weight of the final fully-connected layer of the network. 
$M$ is the number of classes to be predicted.   

The cross-entropy loss is used for the training, which is defined as:
\begin{equation}
	\mathcal{L} = -\mathrm{y}^{*} \cdot \mathrm{log}(\frac{\mathrm{exp}(\mathrm{y})}{\mathrm{\textbf{1}}^\mathrm{T} \mathrm{exp}(\mathrm{y})})
	\label{eq02}
\end{equation}
Here $\mathrm{y}^{*}$ denotes ground-truth in one-hot representation. 
$\mathrm{\textbf{1}}$ is one vector.
The backpropagation is used to train the parameters $\alpha(\cdot)$.
The gradient is defined as:
\begin{equation}
	\frac{\partial{\mathcal{L}}}{\partial{\theta}} = \frac{\partial{\mathcal{L}}}{\partial{\mathrm{y}}}\sum_{n}\frac{\partial{\mathrm{y}}}{\partial{{\alpha}_n}}\frac{\partial{{\alpha}_n}}{\partial{\theta}}=\frac{\partial{\mathcal{L}}}{\partial{\mathrm{y}}}\sum_{n}\mathrm{\textbf{W}}\mathrm{\textbf{f}}_{n}\frac{\partial{{\alpha}_n}}{\partial{\theta}}
	\label{eq03}
\end{equation}
\color{black}

\subsubsection{Local Features' Dimensionality Reduction}

A commonly used feature dimension reduction method \citep{jegou2012negative} is incorporated to reduce the dimension of local features.
We firstly pre-process the local features with L2 normalization.
Then, their dimension is reduced using PCA to generate \mbox{40-dimensional} features.
Finally, the features are processed again through a L2 normalization, as it has been demonstrated by \cite{jegou2012negative} that the re-normalization provides a better mean average precision in image retrieval tasks.

\section{Hierarchical Navigable Small World Graph~Database}
\label{HNSW}

Our system employs the HNSW graph to index the generated global features.
The proposed method is selected as it constitutes a reliable technique that outperforms other contemporary approaches, such as tree-based BoW \citep{muja2014scalable}, product quantization \citep{jegou2011product}, and locality sensitive hashing \citep{andoni2015optimal}.
The following sub-sections describe its properties and the way HNSW is used to construct the graph-based visual database.

\subsection{Hierarchical Navigable Small World}
\begin{figure*}
	\centering
	\includegraphics[width=0.9\textwidth]{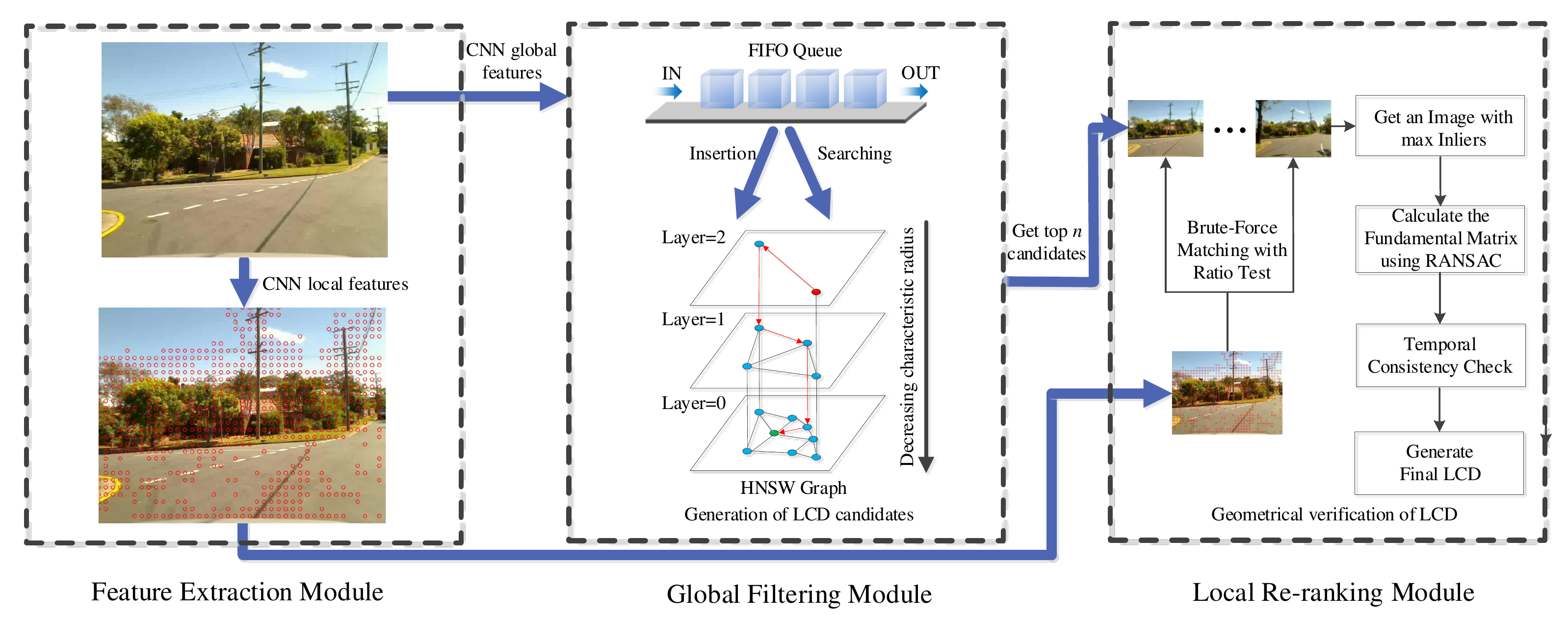}
	\caption{An overview of the proposed loop closure detection pipeline.
		Global and local Convolution Neural Network (CNN) -based features are extracted as the incoming image stream enters the system.
		The global features enter the First-In-First-Out (FIFO) queue, and subsequently, they are fed into the HNSW graph~\citep{malkov2018efficient}, to generate the incremental database.
		Simultaneously, the top $n$ nearest neighbors are retrieved using the global feature, while a brute-force matching technique between the candidate image-pairs is performed at the local features space.
		A ratio test is implemented to eliminate false matches in conjunction with a RANSAC-based geometrical verification check.
		Finally, a temporal consistency check is employed to approve the final loop closure pair.}
	\label{fig:hnsw}
\end{figure*}

HNSW is a fully graph-based incremental $k$-Nearest Neighbor Search ($k$-NNS) structure, as shown in Fig. \ref{fig:hnsw}.
It is based on the Navigable Small World (NSW) model \citep{kleinberg2000navigation}, which follows a logarithmic or polylogarithmic scaling of greedy graph routing.
Such models are important for understanding the underlying mechanisms of real-life networks' formation.

A graph \textcolor{black}{ ${\boldsymbol{G}=(\boldsymbol{V},\boldsymbol{E})}$ } formally consists of a set of nodes (\textit{i.e.,} feature vectors) $\boldsymbol{V}$ and a set of links $\boldsymbol{E}$ between them.
A link $e_{ab}$ connects node $a$ with node $b$ in a directional manner, \textit{i.e.,} form $a$ to $b$, on HNSW.
The neighborhood of $a$ is defined as the set of its immediately connected nodes.
HNSW exploits strategies for explicit selection of the graph's enter-point node, separates links of different length scales, and chooses neighbors using an advanced heuristic.
Then, the search process is performed in a hierarchical multilayer graph, which allows logarithmic scalability.

\subsection{Database Construction}

In BoW-based approaches, the visual vocabulary is usually constructed using $k$-means clustering. A search index is built over the visual words, which are generated using feature descriptors extracted from a training dataset.

HNSW has the property of incremental graph building \citep{malkov2018efficient}.
The image features can be consecutively inserted into the graph structure.
An integer maximum layer $l$  is randomly selected with an exponentially decaying probability distribution for every inserted element.
The insertion process starts from the top layer to the next layer by greedily traversing the graph to find the $ef$ closest neighbors to the inserted element $q$ in the layer.
The found closest neighbors from the previous layer will be used as an enter point to the next layer.
A greedy search algorithm is used to find the closest neighbors in each layer.
The process repeats until the connections of the inserted elements are established on the zero layer.
In each layer higher than zero, the maximum number of connections that an element can have per layer is defined by the parameter $M$, which is the only meaningful construction parameter.
The construction process of the HNSW graph is illustrated in the middle of Fig. \ref{fig:hnsw}.

During the mobile robot's movement, the deep global features of the images are inserted into the graph vocabulary. The whole process is on-line and incremental,  thus eliminating the need for prebuilt vocabulary. Therefore, the use of HNSW ensures the robot's working in various environments.

\subsection{$k$-NN Search}

The $k$-NN Search algorithm in HNSW is roughly equivalent to the insertion algorithm for an item in layer $l=0$.
The difference is that the closest neighbors found at the ground layer are returned as the search result.
The search quality is controlled by the parameter $ef$.

The distance between two global features or nodes in the HNSW graph, indicates the corresponding images' similarity.
We use the normalized scalar product (cosine of the angle between vectors) to compute the distance of two nodes during graph construction and $k$-NN search, which is calculated as: \color{black}
\begin{equation}
s_{pq}=\frac{X_p^T\cdot{X_q}}{{\left \| X_p \right \|}_2 \cdot {\left \| X_q \right \|}_2}.
\label{eq04}
\end{equation}
where $s_{pq}$ is the distance score between images $I_p$, $I_q$ and $X_p$, $X_q$ are the global description vectors.
${\left \| X \right \|}_2=\sqrt{{X^T}X}$ denotes the Euclidean norm of vector $X$.
Since we aim to build a computational inexpensive system, we have chosen to make use of the Advanced Vector Extensions instructions to accelerate the distance computation.

\section{Detection Pipeline}
\label{pipeline}

\subsection{System Overview}

As the robotic platform navigates into the working area, its incoming sensory information, provided by the mounted camera, passes through the CNN to extract the visual features.
Firstly, the global features enter the First-In-First-Out (FIFO) queue, aiming to avoid early visited locations' detection, and then are placed into the database.
The $n$ most similar locations, indicated by $k$-NN, are selected, while an image-to-image correlation eliminates false positive matches through a ratio test.
Eventually, geometrical and temporal consistency checks are employed to generate the final loop closure pair.
An overview of the proposed scheme is illustrated in Fig. \ref{fig:hnsw}, while its steps are described in Algorithm \ref{algorithm1}.

\begin{algorithm}[!t]
  \caption{{\color{black}Our loop closure detection pipeline}}
  \label{algorithm1}
    {\bf Input:} the image $I_i$ captured by the visual sensory module during robot's navigation; the excluded area, defined by frame $N_{non}$ as $\psi \times \phi$, where $\psi$ is a temporal constant and $\phi$ is the frame rate of the camera; the returned number of nearest neighbors $n$; the threshold of inlier points $\tau$. \\
    {\bf Output:} whether the $i$ detection constitute a loop closure or not.

    \begin{algorithmic}[1]
    \State initialize a First In First Out (FIFO) queue $Q$.
    \While {true} $\triangleright$ perform the loop closure detection pipeline during robot's mission.
        \State $I_i$ $\leftarrow$ read the current image.
        \State $X_i, L_i$ $\leftarrow$ extract global and local visual features.
        \If{  ($i > N_{non}$) }
            \State $X_{pre}$ $\leftarrow$ pop the FIFO queue $Q$.
            \State add $X_{pre}$ to HNSW graph visual database.
            \State $k$-nearest neighbor search of $X_i$ in the database to

            \quad obtain the $n$ among them.
            \State $inlier_{max}$ $\leftarrow$ $-1$, $ind$ $\leftarrow$ $-1$

             \For  {$r=1$ to $n$}
                \State perform geometrical verification for $L_i$ and $L_{r}$
                \If{failed}
                    \State continue
                \EndIf
                \State $inlier$ $\leftarrow$ the number of inliers
                \If{ $inlier > inlier_{max}$}
                     \State $inlier_{max}$ $\leftarrow$ $inlier$
                     \State $ind$ $\leftarrow$ $r$
                \EndIf
             \EndFor

             \State temporal consistency check for $L_i$ and $L_{ind}$
             \If{success}
             \State Loop detected.
             \EndIf
        \EndIf
        \State push $X_i$ to the FIFO queue $Q$.
    \EndWhile
  \end{algorithmic}
\end{algorithm}

\subsection{Retrieval Strategy}
The $n$ most similar locations are determined via the HNSW's $k$-NN search using the query's extracted global feature.
Since the image frames are captured sequentially, the adjacent locations to query, \textit{i.e.,} images acquired in close time proximity, are highly possible to share semantic information yielding to high similarities among them.
When searching the database this area should be avoided, so as to keep the system safe from false positive detections.
{\color{black}
Therefore, we use the FIFO queue to store images' global representations.
As shown in Algorithm \ref{algorithm1}, the global feature $X_i$, belonging to image $I_i$, firstly enters the queue $Q$, and subsequently it remains there aiming to be inserted at the HNSW graph when the robot runs out of the non-search area.
The non-search area is defined based on a temporal constant $\psi$, and the camera's frame rate $\phi$.
Consequently, when we use the current feature as query, it will only search in database area defined via \mbox{$N - \psi \times \phi$}, where $N$ is the number of the entire set of camera measurements up to time $i$.
As a final note, the images in the non-search area will never appear in the results.
\color{black}}

\begin{table*}[t!]
\caption{Descriptions of the Used Datasets}
\label{table_dataset}
\begin{center}
\renewcommand\tabcolsep{2pt}
\resizebox{\textwidth}{!}{
\begin{tabular}{cc|c|c|c|c|c}
\toprule
Dataset & & Description  & Image Resolution (px) & \# Images & Frame rate ($Hz$) & Distance ($km$) \\
\midrule
KITTI \citep{Geiger2012CVPR} & Seq\# 00 & Outdoor, dynamic  & $1241\times376$ & 4541 & 10 & 3.7  \\
& Seq\# 02 &    & $1241\times376$ & 4661 &  & 5.0 \\
& Seq\# 05 &    & $1226\times370$ & 2761 &  & 2.2 \\
& Seq\# 06 &    & $1226\times370$ & 1101 &  & 1.2 \\
\midrule
Oxford & New College \citep{smith2009new} & Outdoor, dynamic  & $512\times384$ & 52480 & 20 & 2.2 \\
& City Center \citep{cummins2008fab} &   & $640\times480$ & 1237 &  10 & 1.9 \\
\midrule
Malaga 2009 \citep{blanco2009collection} & Parking 6L  & Outdoor, slightly dynamic  & $1024\times768$ & 3474 & 7 & 1.2\\
\midrule
St. Lucia \citep{glover2010fab} & 100909 (12:10) & Outdoor, dynamic  & $640\times480$ & 19251 & 15 & $\sim$17.6 \\
 & 100909 (14:10) &  &  & 20894 &  & \\
 & 180809 (15:45) &  &  & 21434 &  & \\
 & 190809 (08:45) &  &  & 21815 &  & \\
\bottomrule
\end{tabular}
}
\end{center}
\end{table*}

\subsection{Image-to-Image Matching}
As described in Section~\ref{ImageRepresentation}, we extract local deep features for each incoming image.
Thus, the matching process is performed between the query $q$ and the $n$ closest neighbors based on a brute-force matching algorithm.
This technique is rarely reported in the literature for visual data association due to the presented high complexity. 
However, when low-dimensional floating point global descriptors are used, such in our case, brute-force matching does not demand relatively high computation time.
At last, a distance ratio check~\citep{lowe2004distinctive}, defined through a threshold $\varepsilon$, is employed on the proposed pair.

\subsection{Geometrical Verification}

Our system incorporates a geometrical verification step to discard outliers, \textit{i.e.,} false positive detections.
In order to achieve this, we compute the fundamental matrix $T$ between the chosen candidate pair of images using a RANdom SAmple Consensus (RANSAC) -based scheme \citep{torr1997development}.
We record the candidate with the highest number of inliers when the calculation succeeds.

\subsection{Temporal Consistency Check}

As a final step, a temporal consistency check is employed intending to examine whether the aforementioned conditions are met for $\beta$ consecutive camera measurements similarly to \cite{tsintotas2018assigning}.
This way, the proposed pipeline may lose a possible loop closing identification in cases where the query image is the initial in a sequence of pre-visited locations, however we prefer to prevent the system from wrong identifications preserving.
When the aforementioned conditions are met, the matched pair is recognized as a loop closure event.

\section{Benchmark}
\label{Experimental Protocol}

\subsection{Benchmark Datasets}
\label{BenchmarkDatasets}

Eleven challenging and publicly available image-sequences have been chosen to evaluate the performance of our framework.
These datasets are captured in different operating environments, \textit{e.g.,} various lighting conditions, strong visual repetition and dynamic occlusions such as cars and pedestrians.
A detailed description of each image-sequence is listed in Table~\ref{table_dataset}.
Regarding KITTI vision suite \citep{Geiger2012CVPR}, Malaga 2009 Parking 6L (Malaga6L) \citep{blanco2009collection}, and St. Lucia \citep{glover2010fab}, the incoming visual stream is obtained via a camera mounted on a moving car, while New College \citep{smith2009new}, and City Center \citep{cummins2008fab} are recorded through the vision system of a wheeled robot.
Malaga6L, New College and City Center are composed of stereo images; in the course of our experiments, we have chosen the left camera stream for the first and the right camera for the rest.
Our experimental setup is chosen according to the work of \cite{kazmi2019detecting}.

\subsection{Ground Truth Labeling}
\begin{figure}[t!]
	\centering
	\includegraphics[width=0.5\textwidth]{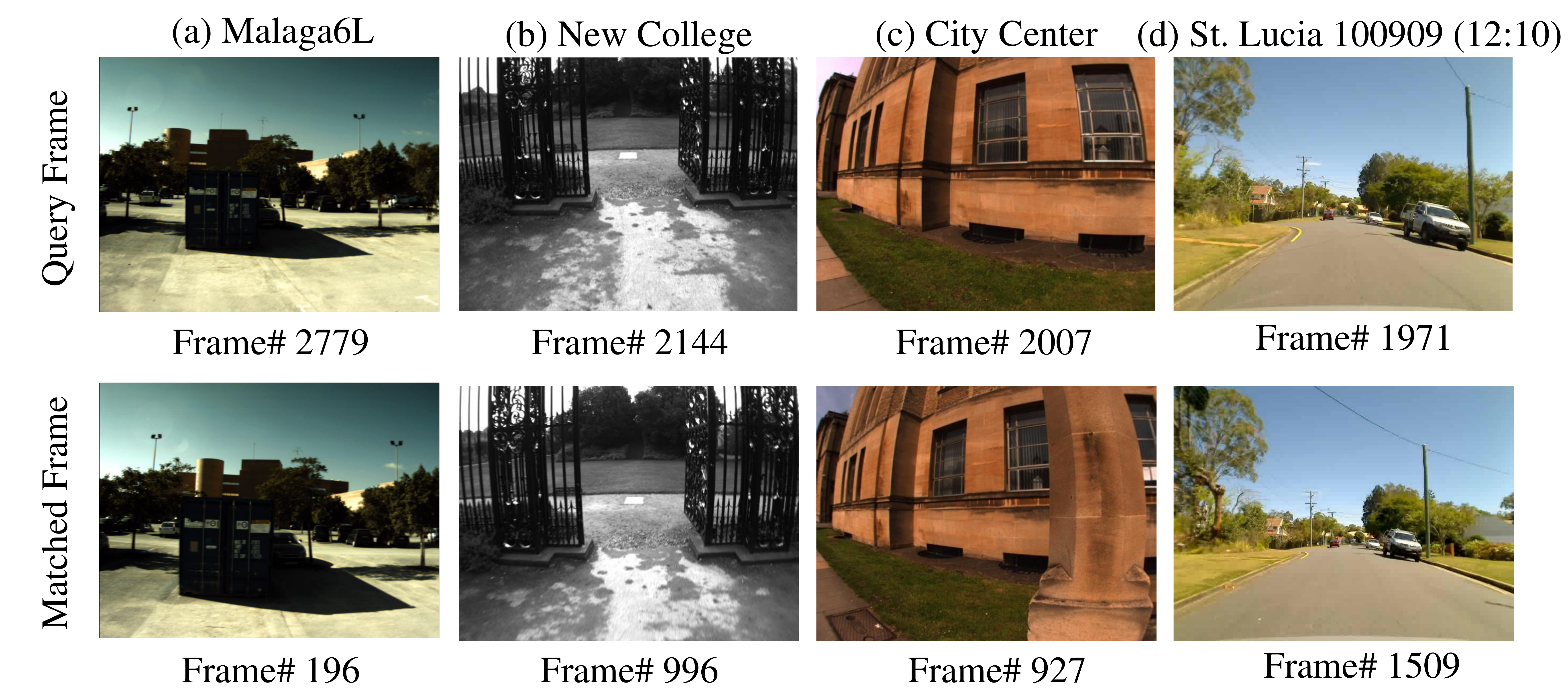}
	\caption{Examples of image pairs which are not correctly labeled in the ground truth data derived via the Global Positioning System (GPS) logs.}
	\label{fig:gt_noloop}
\end{figure}
\begin{figure*}
\centering
 \includegraphics[width=1.0\textwidth]{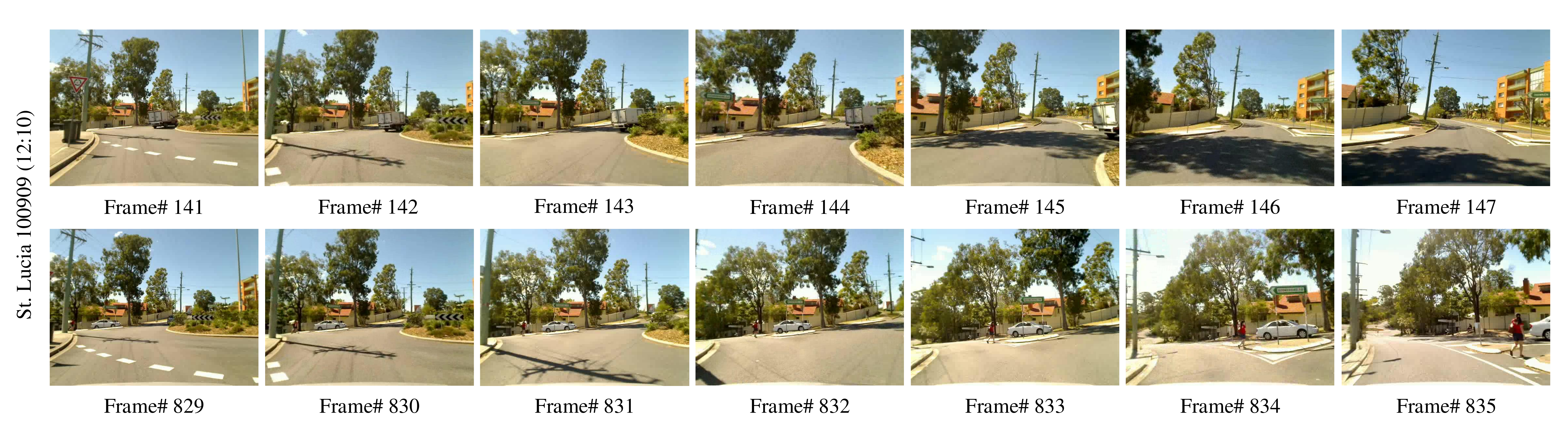}
  \caption{
  A labeling correction: the image sequence in the first row shows the robot's trajectory as it turns to the right road, while in the second row, it turns to the left road at the same place.
  Frames \#835 and \#147 are visually different but are labeled as loops according to the GPS, for its distance is lower than 10 $m$.
  During our experiments, these images are considered as true negative pairs.
  }
\label{fig:samepos_difvis}
\end{figure*}

Commonly, the ground truth data, referring to the correct loop events, is generated according to the Global Positioning System (GPS) logs.
For example, St. Lucia and Malaga6L utilize a GPS distance-range of 10 $m$ and 4 $m$ from the query, respectively, to define the ground truth.
We carefully checked this data for each dataset recognizing that some image pairs were not accurately labeled, as shown in Fig.~\ref{fig:gt_noloop}.
In many cases, this occurs owing to the robot traversing through locations that surpass the GPS's distance threshold, though the captured visual content might be similar.
However, in such cases, if a valid fundamental matrix is computed, the transformation matrix between the two camera poses can be available. 
Such pairs should be treated as true positive loop closure events.
Another problem concerns the situation wherein the robot's viewpoint differs from the viewpoint confronted in its first traversal.
Regardless of the system being precisely located at the same place, these image pairs are considered true negative events.
An exemplar case of this situation is illustrated in Fig.~\ref{fig:samepos_difvis}.


Considering the GPS logs are not accurate, we adopt human labelling for the ground truth generation.
We produce image pairs which are located less than 40 meters in GPS logs.
Then these pairs are labelled by asking whether they are from the same place by crowdsourcing.
During labeling, if a decision was hard to be taken, the proposed pairs are re-checked by experts familiar with the place recognition task.
Each of the aforementioned datasets is processed two times before used, while for the KITTI image-sequences, the data were accurate enough avoiding this procedure.
Our accurate, manually-labeled ground truth files are made publicly available in order to facilitate further studies.

\section{Experimental Results}
\label{results}

This section presents the experiments conducted to demonstrate the proposed pipeline's effectiveness.
Our setup including training strategy, parameters and evaluation metrics are introduced in \ref{sec_settings}, while different settings for the proposed features' extraction module 
\color{black}
are evaluated in \ref{sec_scales}.
Next, we analyze the HNSW parameterization in \ref{sec_hnsw}, and evaluate the geometrical verification process in \ref{sec_gc}. 
A comparison of our global feature with two contemporary CNN-based features is presented in \ref{sec_gf}.
The system's performance and quantitative comparison with the state-of-the-art are presented in \ref{sec_performance}.
Finally, we measure our system’s complexity on the representative datasets in \ref{sec_time}.
\color{black}

\subsection{Experimental Settings}
\label{sec_settings}

\color{black}
\subsubsection{Training Strategy}

Since our feature extractor is hard to get trained directly, owing to the employment of the attention module, a two-step strategy is applied.
Firstly, our base network is trained, leaving the attention module out; subsequently, two Fully Connected Layers (FCLs) are adjoined for the classification.
ResNet50, upon which the proposed system is built, is trained on the ImageNet \citep{russakovsky2015imagenet}
and then the model is fine-tuned on a large-scale landmark dataset \citep{weyand2020google}.
The cross-entropy loss is used for the image classification.

Next, when the base network is trained, its weights are squeezed.
The attention module is added, and the resulted score map is used to pool the features by a weighted sum.
Subsequently, the features enter the fully connected layer for the classification with the cross-entropy loss.
Finally, we use this model to obtain discriminative deep features.
\color{black}

\subsubsection{Training Parameters}

Our network was trained through the stochastic gradient descend (SGD) optimizer.
An initial learning rate of 0.001 and 25 epochs as the maximum number for training was selected, with its rate being halved every 10 epochs.
Similarly, the same optimizer was chosen for the attention module with an initial learning rate set at 0.01 at the maximum number of 20 epochs, while the learning rate is halved every 10 epochs.
We implemented the two networks using the batch size of 256.

\subsubsection{Baseline Approaches}

The compared methods include classic and recently published place recognition systems namely: DLoopDetector \citep{galvez2012bags}, \cite{tsintotas2018assigning}, PREVIeW \citep{bampis2018fast}, iBoW-LCD \citep{garcia2018ibow}, Kazmi et al. \citep{kazmi2019detecting}, as well as our previous method {\color{black}{\citep{anshan2019}}}.
Most of the chosen methods are implemented using the respective open-source codes.
For Kazmi's method, we directly report their results as published in their article.

\subsubsection{Evaluation Metrics}

\color{black}
For the loop closure detection task, the commonly used metric is the recall rate at 100\% precision. The precision-recall metric is defined as:
\begin{equation}
\textrm{Precision} = \frac{\textrm{true positives}}{\textrm{true positives + false positives}}
\label{eq321}
\end{equation}

\begin{equation}
\textrm{Recall} = \frac{\textrm{true positives}}{\textrm{true positives + false negatives}}
\label{eq322}
\end{equation}
\noindent where true positives is the number of correct identifications, indicating the detected loop closures are true loops according to the ground truth.
False positives is the number of wrong detections, representing the identifications found by the algorithm; however, these are not labeled to ground truth.
False negatives indicate the number of true loop closure events, which are not found by the algorithm.
\color{black}

\begin{table*}
	\caption{The recall at 100\% precision and the feature extraction speed (ms) on different scales of global and local features. }
	\label{tab:scales}
	\centering
	\scriptsize
	\renewcommand\tabcolsep{5pt}
	\begin{tabular}{c|c|c|c|c|c|c|c|c|c|c|c|c|c|c}
		\toprule
		\multirow{3}{*}{\textbf{Scales (Global)}} & \multicolumn{14}{c}{\textbf{Scales (Local)}}                                                                                                                                       \\
		\cline{2-15}
		& \multicolumn{2}{c|}{\textbf{0.25}} & \multicolumn{2}{c|}{\textbf{0.35}} & \multicolumn{2}{c|}{\textbf{0.5}} & \multicolumn{2}{c|}{\textbf{0.7}} & \multicolumn{2}{c|}{\textbf{1.0}} & \multicolumn{2}{c|}{\textbf{1.4}} & \multicolumn{2}{c}{\textbf{2.0}}  \\
		\cline{2-15}
		& recall & speed             & recall & speed            & recall & speed           & recall & speed           & recall & speed         & recall & speed           & recall & speed          \\
		\toprule
		\textbf{0.25}                                           & 0.9123 & 8.11             & 0.9073 & 8.90              & 0.9010  & 10.01           & 0.9123 & 12.54           & 0.9135 & 16.49         & 0.9261 & 28.25           & 0.9236 & 56.07          \\
		\hline
		\textbf{0.35}                                             & 0.8972 & 8.73             & 0.9273 & 9.40              & 0.9110  & 10.56           & 0.9110  & 13.19           & 0.8960  & 17.14         & 0.8960  & 28.86           & 0.9023 & 56.28          \\
		\hline
		\textbf{0.5}                                              & 0.8972 & 9.61             & 0.9110  & 10.28            & 0.9098 & 11.23           & 0.9261 & 14.00              & 0.9110  & 17.93         & \color{blue}0.9492 & 29.62           & 0.9480 & 57.15          \\
		\hline
		\textbf{0.7}                                              & 0.8972 & 11.73            & 0.9110  & 12.41            & 0.8997 & 13.56           & 0.9248 & 16.09           & 0.9098 & 19.90          & \color{blue}0.9492 & 31.79           & \color{blue}0.9492 & 59.01          \\
		\hline
		\textbf{1.0}                                                & 0.8910  & 14.61            & 0.8985 & 15.25            & 0.8935 & 16.22           & 0.9261 & 18.77           & 0.9035 & 22.64         & 0.9211 & 34.53           & 0.9323 & 61.90           \\
		\hline
		\textbf{1.4}                                              & 0.8922 & 20.95            & 0.9035 & 21.69            & 0.8935 & 22.47           & 0.9286 & 24.96           & 0.9048 & 28.89         & 0.9223 & 40.26           & 0.9336 & 68.97          \\
		\hline
		\textbf{2.0}                                                & 0.8947 & 35.08            & 0.9023 & 35.57            & 0.8960  & 36.46           & 0.9261 & 38.95           & 0.9060  & 42.76         & 0.9223 & 53.97           & 0.9336 & 81.00             \\
		\hline
	\end{tabular}
\end{table*}

\subsubsection{Implementation}

Experiments were performed on a Linux machine with an Intel Xeon CPU E5-2640 v3 ($2.60$ $GHz$) and an NVIDIA {\color{black}Tesla} P40 GPU.
More specifically, only feature extraction was performed on the GPU; any other operation ran on the CPU.
The proposed network is implemented via TensorFlow, yet bindings are provided in C++.
Besides, to test the speed on embedded devices, we additionally implemented FILD++ on an NVIDIA Jetson TX2 GPU and report the respective outcome in subsection~\ref{sec_time}.

\color{black}
\subsection{Image Scales Evaluation}
\label{sec_scales}

The original DELF utilizes image pyramids to generate descriptors of different scales.
It uses 7 different scales ranging from 0.25 to 2.0, which are a $\sqrt{2}$ factor apart.
As processing times are crucial for mobile robotic applications, we propose to use only one scale for global feature extraction and another one scale for local feature extraction.

We conduct extensive experiments to evaluate the recall and the extraction speed of using different feature extraction scales.
For KITTI 00 dataset, the results of different combination of scales for extracting global and local features are given in Table~\ref{tab:scales}.
\color{black}
The 3 highest recall scores are marked in blue.
As shown, the scales of 0.7 and 2.0 for global and local features, respectively, reach the highest recall rate at 100\% precision.
A similar score is obtained at the scales of 0.5 and 0.7 for global features and 1.4 for local features.
However, considering the extraction time, we chose the scales of 0.5 and 1.4 which achieve the same recall through a timing below 30 $ms$.
It is also notable that for the scales of 0.25 for both global and local deep features, the extraction time is only 8.11 $ms$.
Our algorithm's parameters are summarized in Table~\ref{table_parameter}, determined via the experimentation reported in Sections~\ref{sec_scales}, \ref{sec_hnsw}, \ref{sec_gc}.

\begin{table}[!b]
	\caption{Parameters list}
	\label{table_parameter}
	\begin{center}
		\resizebox{\columnwidth}{!}{
			\begin{tabular}{c||c}
				\hline
				Image scale for global feature extraction $s_g$ & 0.5 \\
				\hline
				Image scale for local feature extraction $s_l$ & 1.4 \\
				\hline
				Score threshold of local feature $\delta$ & 15 \\
				\hline
				Number of nearest to $q$ elements to return, $ef$ & 40 \\
				\hline
				Maximum number of connections for each element per layer, $M$ & 48 \\
				\hline
				Search area time constant, $\psi$ & 40 \\
				\hline
				Ratio of ratio test, $\varepsilon$ & 0.7 \\
				\hline
				Images temporal consistency, $\beta$ & 2 \\
				\hline
				Number of of matches for geometrical verification, $n$ & 5 \\
				\hline
			\end{tabular}
		}
	\end{center}
\end{table}

\begin{figure}[t]
	\begin{center}
		\begin{multicols}{2}
			\includegraphics[width=1.0\linewidth]{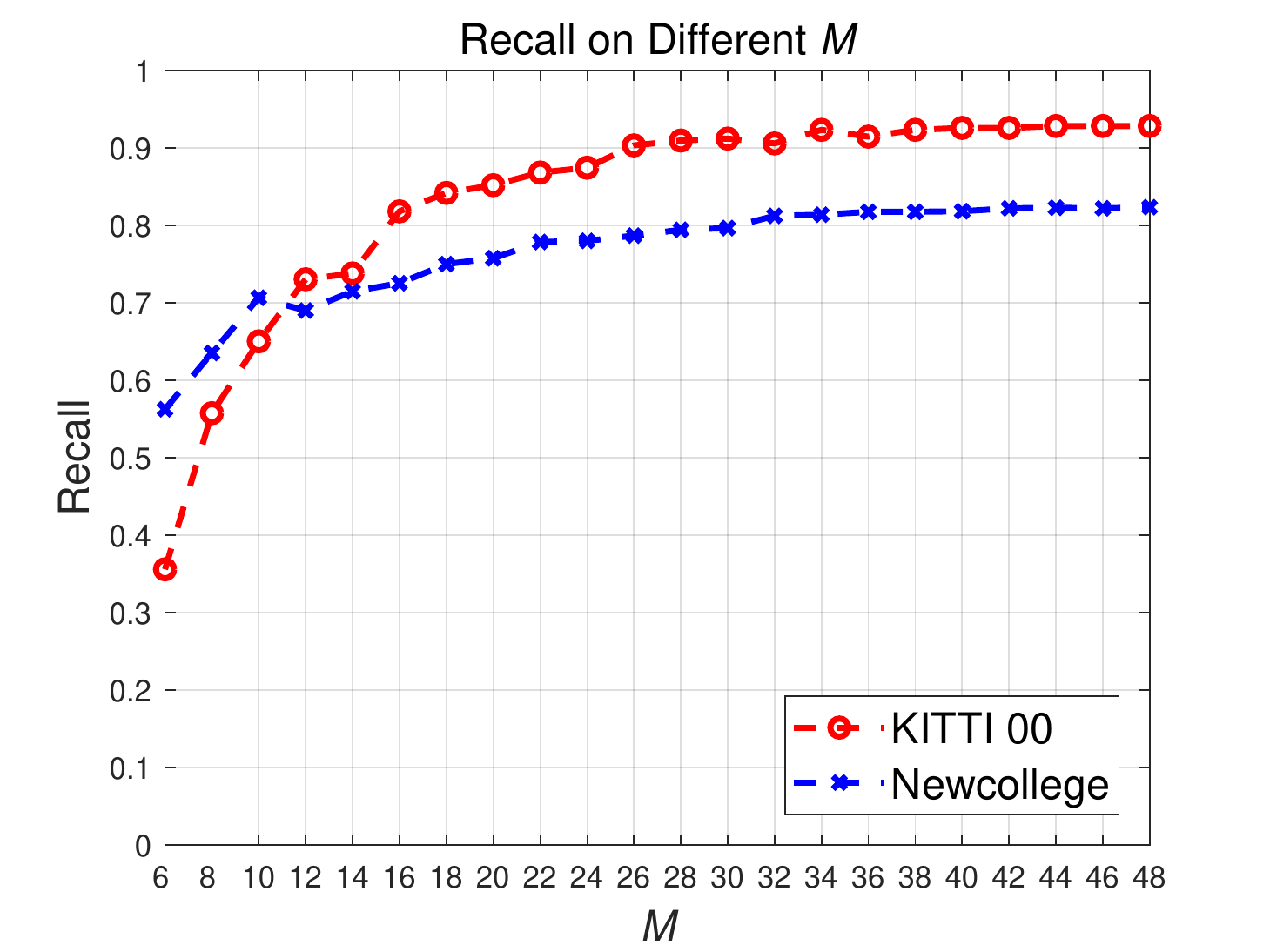} \par
			\includegraphics[width=1.0\linewidth]{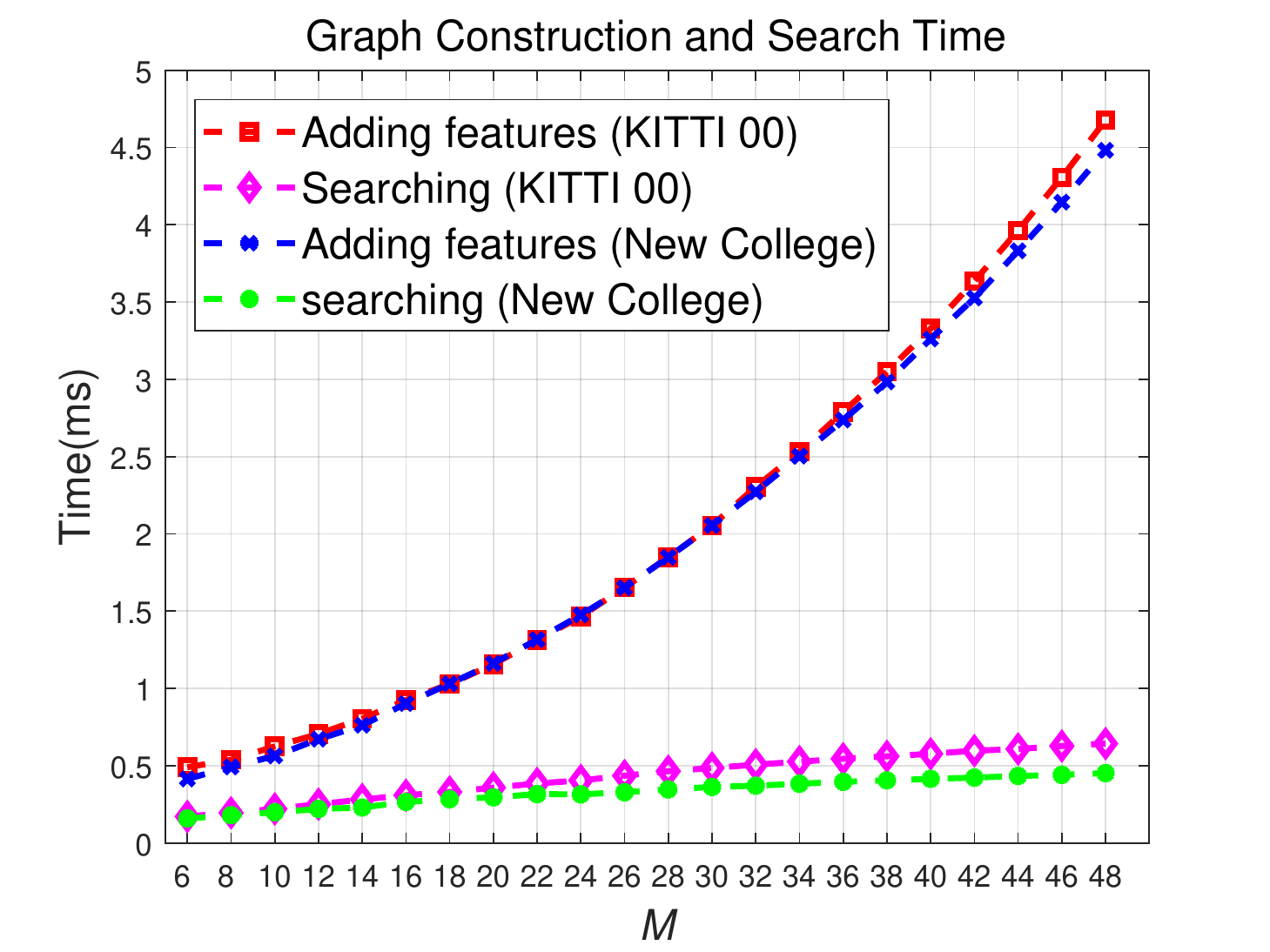} \par
		\end{multicols}
	\end{center}
	\caption{Evaluating the parameter $M$ on KITTI 00 \citep{Geiger2012CVPR} and New College \citep{smith2009new}.
		(Left) Our pipeline's recall scores for perfect precision using a variety of values ranging from 6 to 48.
		(Right) The timing needed for new feature addition and database search.}
	\label{fig:eval_M}
\end{figure}

\begin{figure}[t]
	\begin{center}
	\begin{multicols}{2}
   	\includegraphics[width=1.0\linewidth]{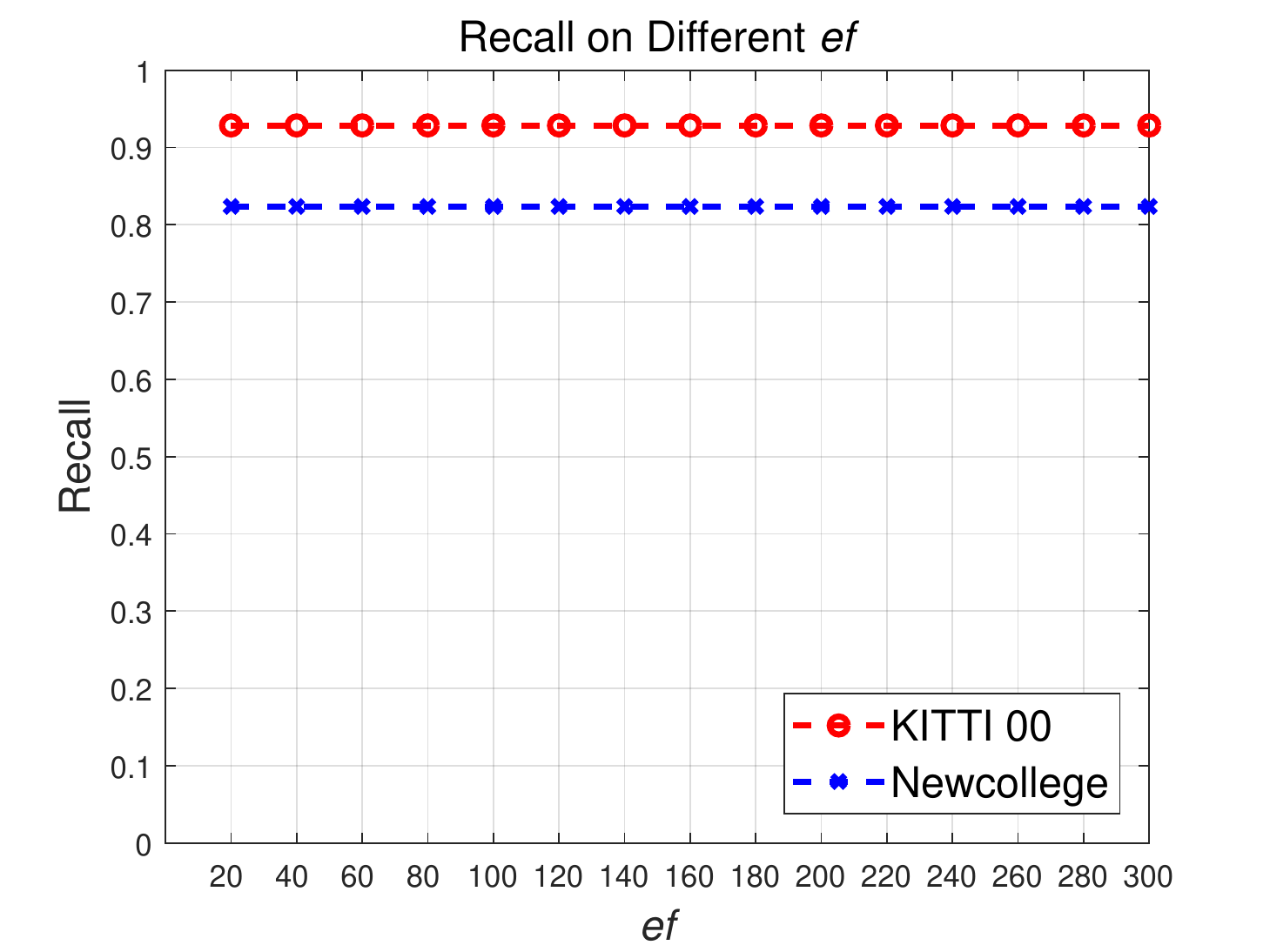} \par
   	\includegraphics[width=1.0\linewidth]{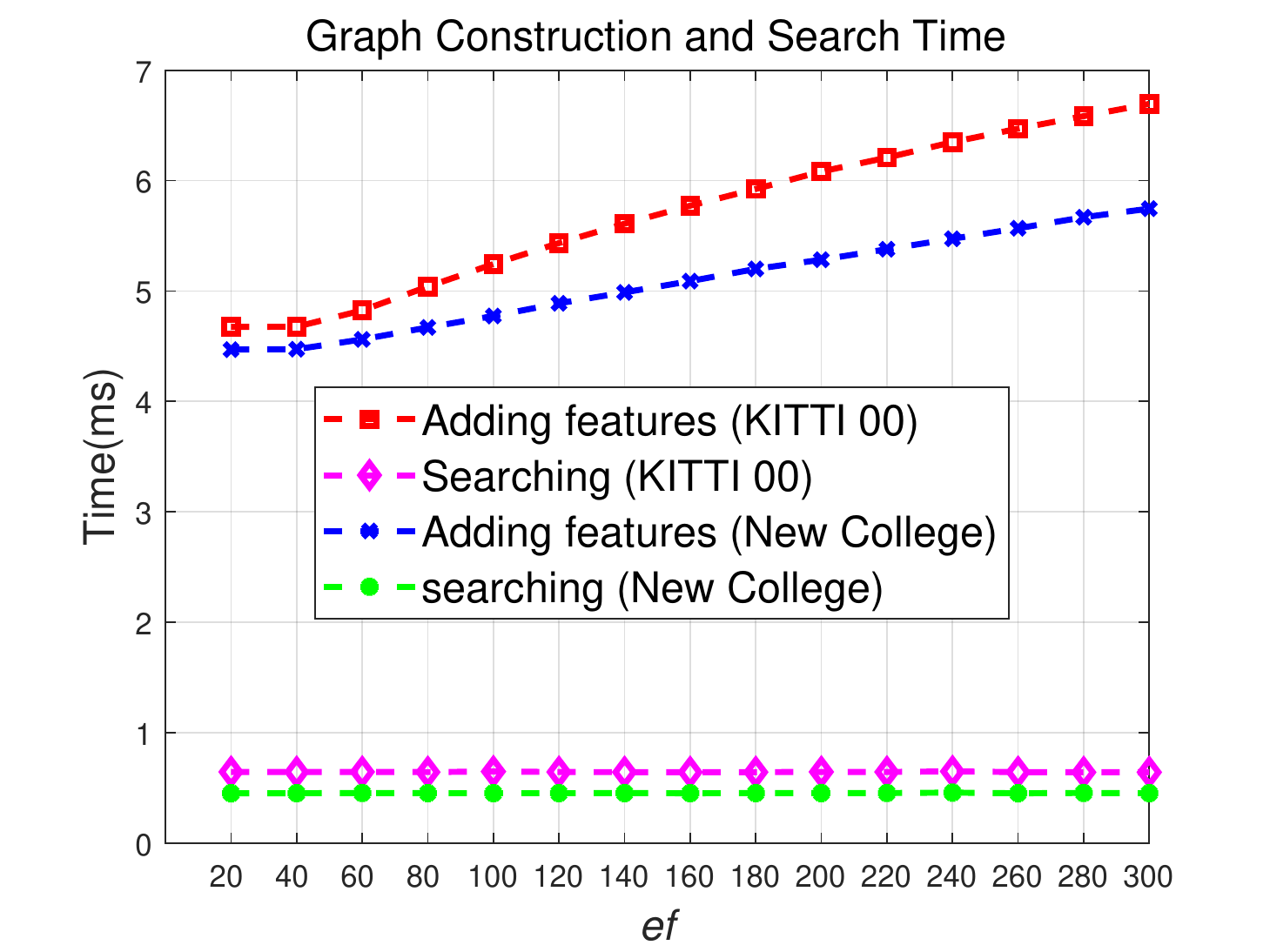} \par
 	\end{multicols}
	\end{center}
   \caption{Evaluating the parameter $ef$ on KITTI 00 \citep{Geiger2012CVPR} and New College \citep{smith2009new}.
   (Left) Our pipeline's recall scores for perfect precision using a variety of values ranging from 20 to 300.
   (Right) The timing needed for new feature addition and database search.}
	\label{fig:eval_EF}
\end{figure}

\color{black}
\subsection{Hierarchical Navigable Small World Parameters' Evaluation}
\label{sec_hnsw}
For HNSW graph construction and searching, there are two parameters that could affect the search quality: the number of nearest to $q$ elements to return, $ef$;  and the maximum number of connections for each element per layer, $M$.  The range of the parameter $ef$ should be within 300, because the increase in $ef$  will lead to little extra performance but in exchange, significantly longer construction time. The range of the parameter $M$ should be 5 to 48 \citep{malkov2018efficient} .  The experiments in \citep{malkov2018efficient} show that a bigger $M$ is better for high recall and high dimensional data, which also defines the memory consumption of the algorithm.

We perform the experiments on the KITTI 00 and the New College datasets to choose $M$ and $ef$ for the HNSW graph. 
The parameter $ef$ is set to 40 when we change $M$. 
The number of matches for geometrical verification $n$ is set to 5. As 100\% precision can be reached with the temporal consistency check. The recalls are shown in the left part of Fig.~\ref{fig:eval_M}. 
We can see when $M$ increases, the recall will also increase. 
In the right part of Fig.~\ref{fig:eval_M}, the feature adding time and searching time will be increased when $M$ increases. 
To achieve a better recall, we choose $M=48$ in the following experiments.

For evaluating different $ef$, it can be seen that in the left part of Fig.~\ref{fig:eval_EF}, the recall does not significantly change when the $ef$ increases. 
In the right part of Fig.~\ref{fig:eval_EF}, the feature adding time will be increased when $ef$ increases, while the searching time remains with no growth. 
Therefore, $ef=20$ was selected.

\begin{figure}[ht]
\centering
 \includegraphics[width=0.4\textwidth]{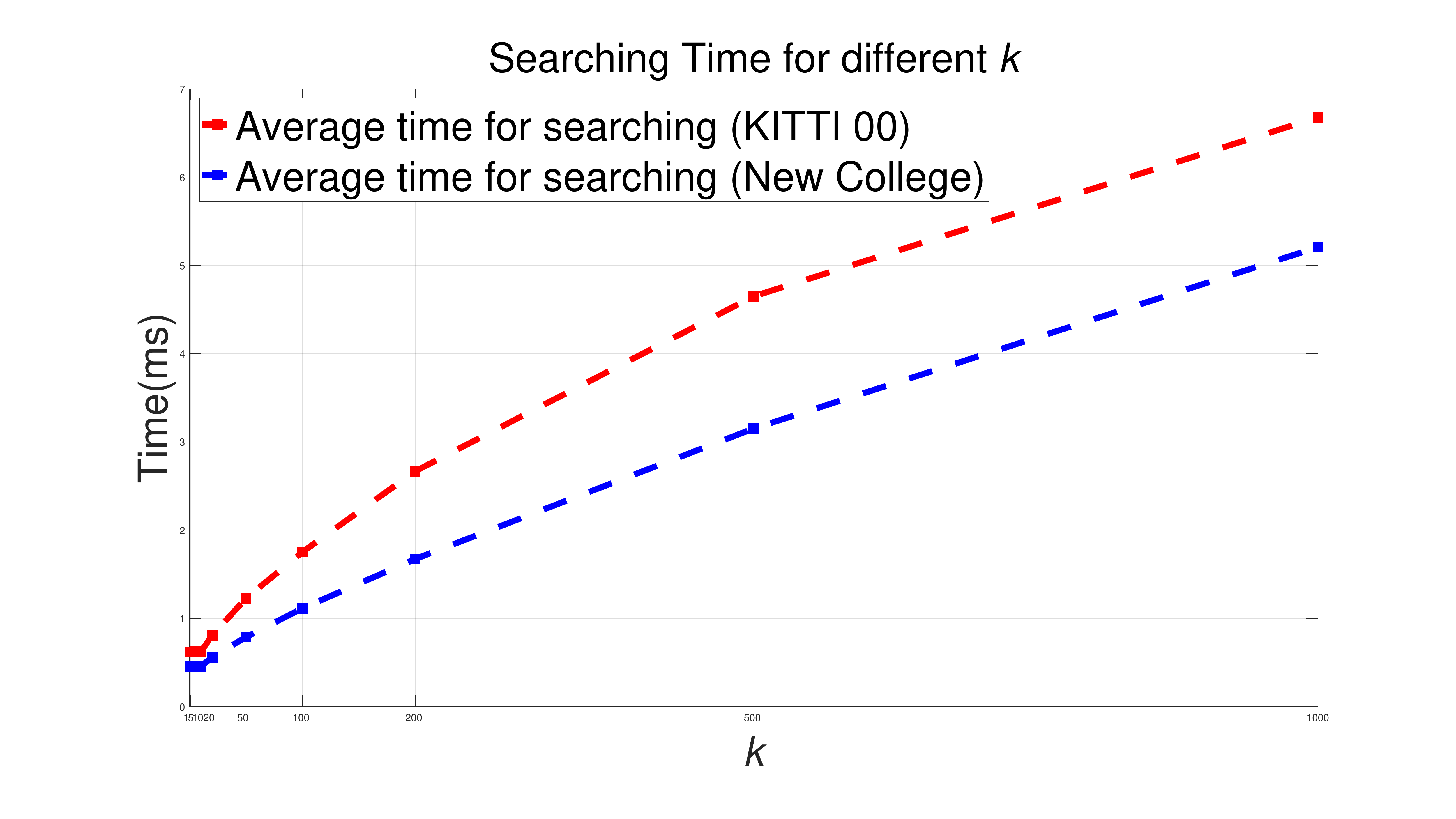}
  \caption{The searching time for different $k$ on the KITTI 00 dataset \citep{Geiger2012CVPR} and the New College dataset \citep{smith2009new}. }
\label{fig:eval_K}
\end{figure}

Besides, we evaluate the searching time of the HNSW graph for different returned number $k$ of nearest neighbors.
As shown in Fig.~\ref{fig:eval_K}, we can see that the searching method costs nearly logarithmic time when increase the returned nearest neighbors.
The time cost accords with the time complexity of the HNSW graph \citep{malkov2018efficient}.
\color{black}

\begin{figure}[t]
	\begin{center}
	\begin{multicols}{2}
   	\includegraphics[width=1.0\linewidth]{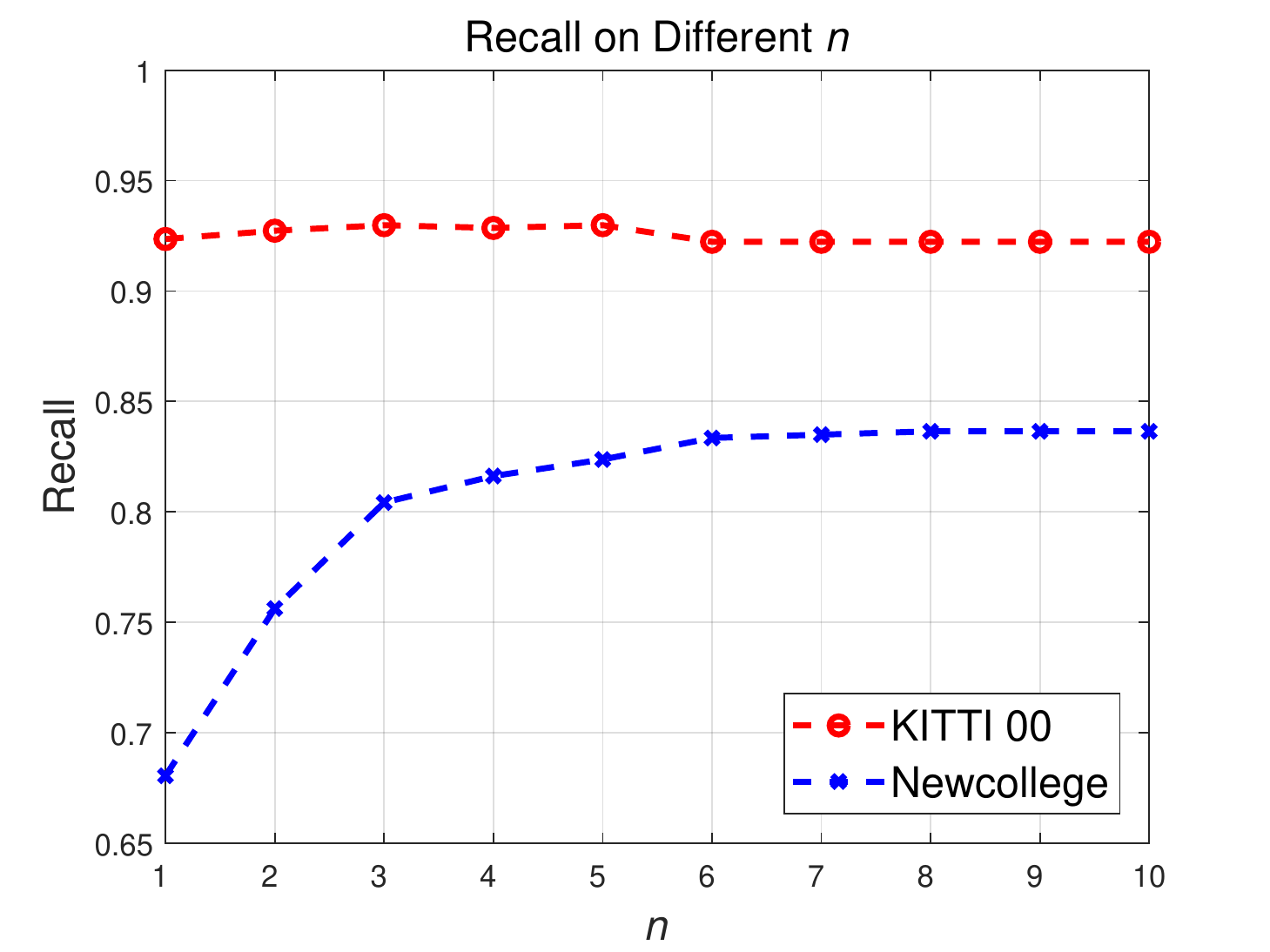} \par
   	\includegraphics[width=1.0\linewidth]{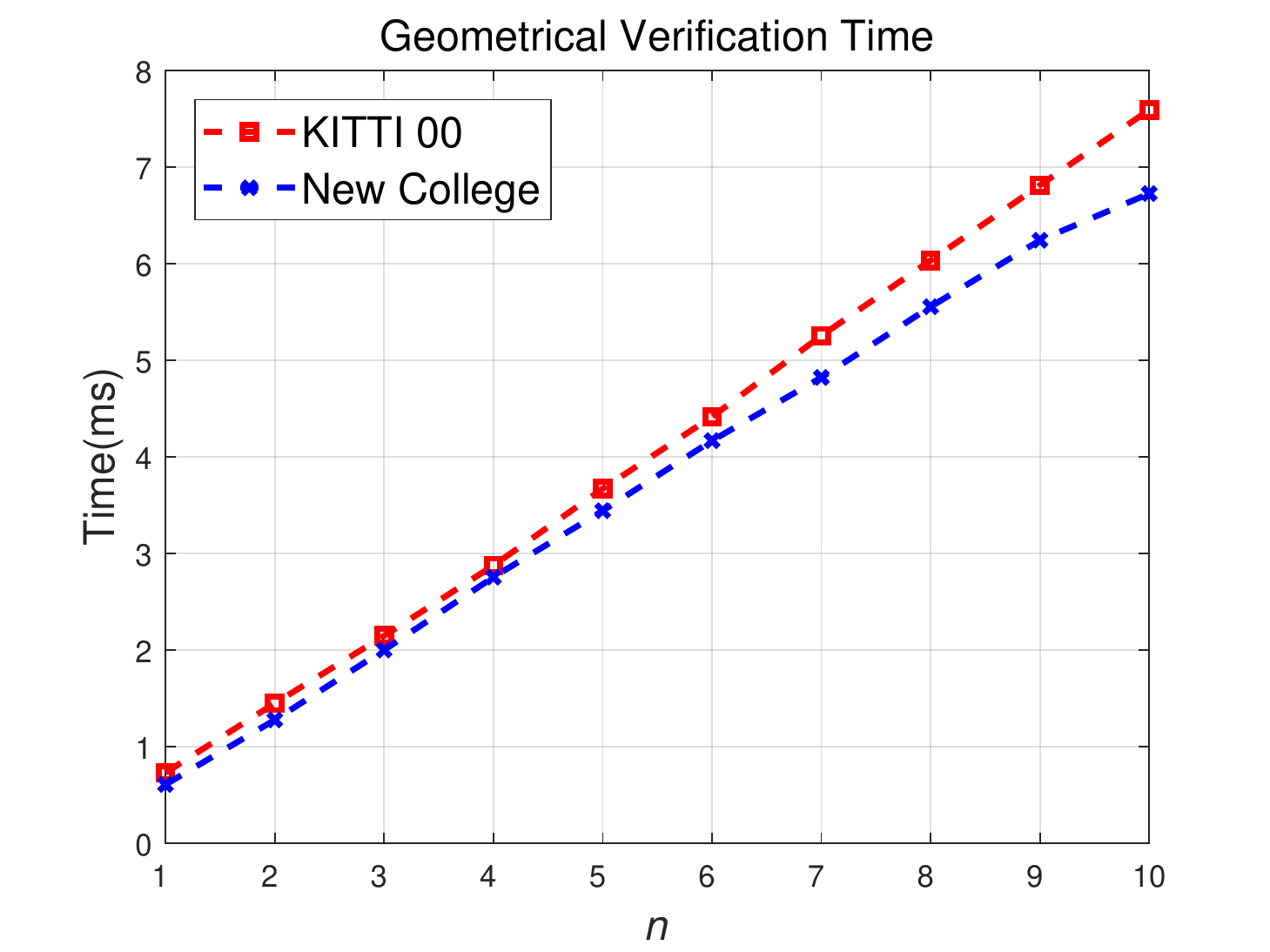} \par
	\end{multicols}
	\end{center}
   	\caption{Evaluating the parameter $n$ on KITTI 00 \citep{Geiger2012CVPR} and New College \citep{smith2009new}.
   	(Left) Our pipeline's recall scores for 100\% precision using a variety of values $n$.
   	(Right) The timing needed for geometrical verification.}
	\label{fig:eval_GV}
\end{figure}

\begin{figure}[t]
	\begin{center}
	\begin{multicols}{2}
   	\includegraphics[width=1.0\linewidth]{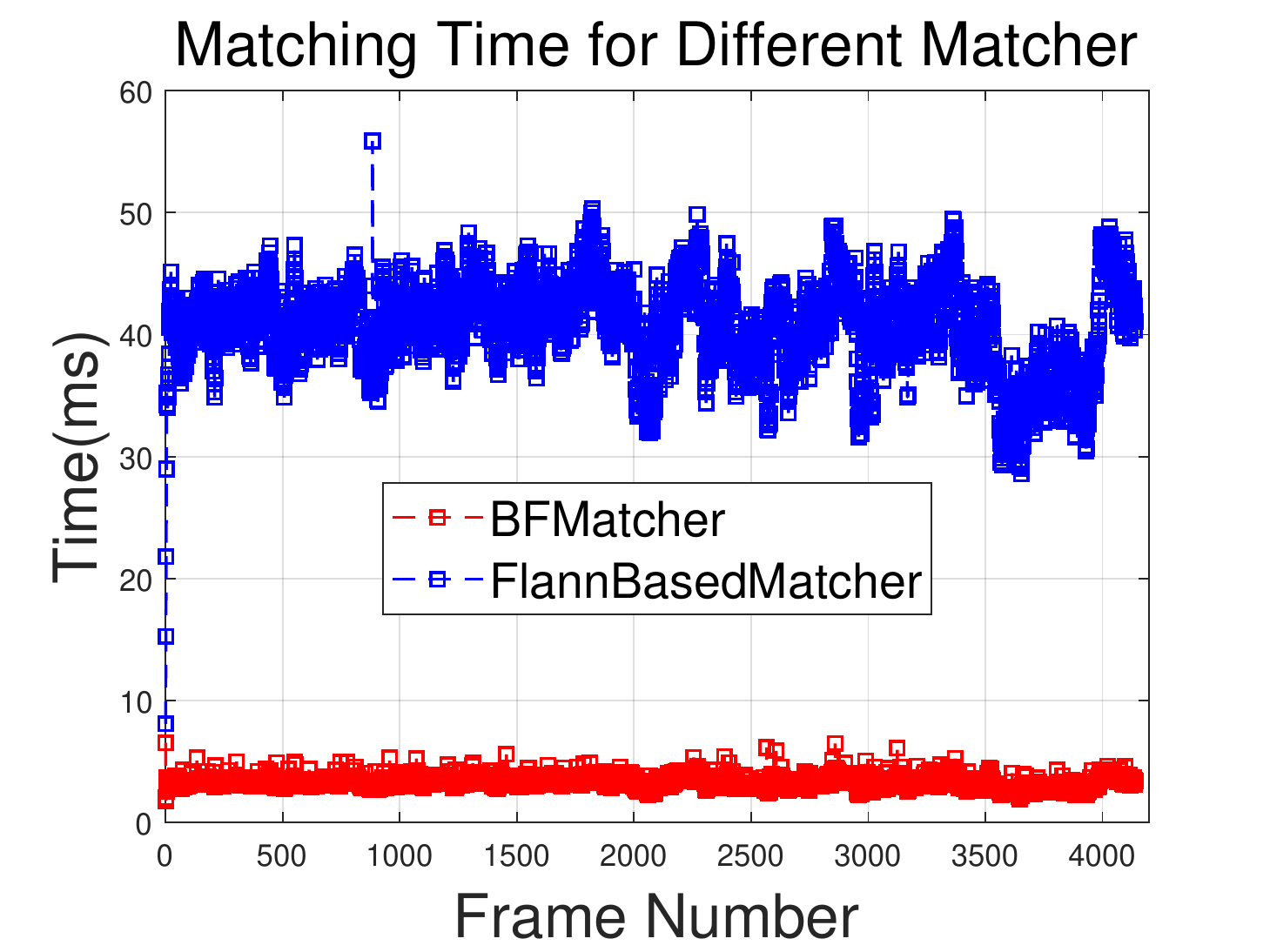} \par
   	\includegraphics[width=1.0\linewidth]{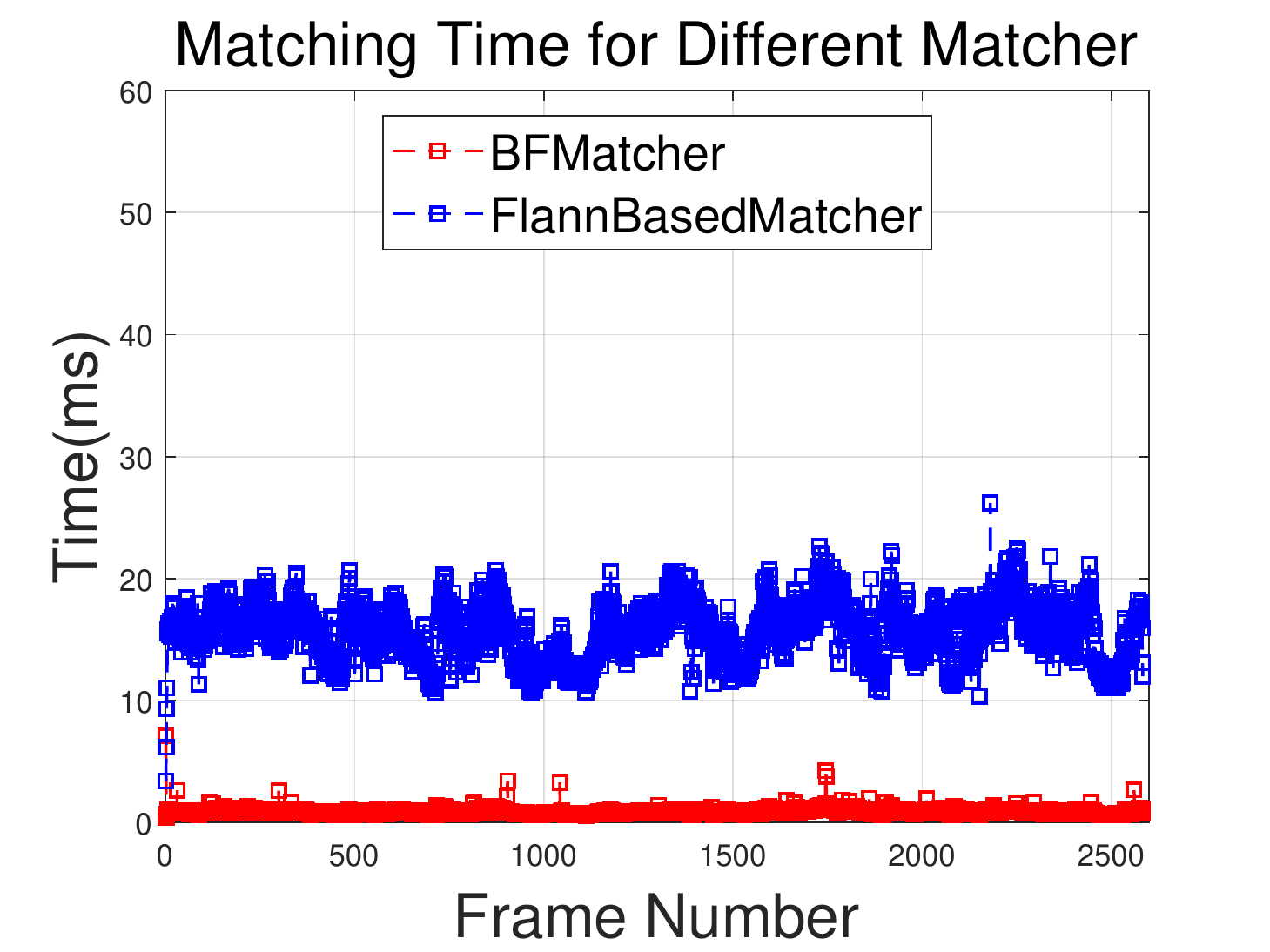} \par
 	\end{multicols}
	\end{center}
   \caption{The image matching time of our algorithm on the KITTI 00 dataset \citep{Geiger2012CVPR} (Left) and the New College dataset \citep{smith2009new} (Right) using different matching strategies.}
	\label{fig:eval_matching}
\end{figure}

\subsection{Evaluating Geometrical Verification}
\label{sec_gc}

As image-to-image matching through RANSAC is computationally costly, we evaluate the parameter $n$ using values ranging from $1$ to $10$.
As shown in Fig.~\ref{fig:eval_GV}, the timing needed for geometrical verification increases linearly with $n$,
as a new RANSAC estimation needs to be done on each round.
For KITTI 00, timing varies from 0.73 $ms$ to 7.59 $ms$ for $n = [1, 2, 3,..., 10]$.
New College timing varies from 0.61 $ms$ to 6.72 $ms$.
However, the higher the value of $n$ the better the performance.
Aiming to achieve a trade-off between recall and computational complexity, we have chosen $n=5$.
\color{black}
We empirically fix the ratio test $\varepsilon$ to 0.7. 
This value is frequently used for image matching using SIFT \citep{lowe2004distinctive} and SURF \citep{bay2006surf}.
\color{black}

Furthermore, we evaluate the processing time for two different {\color{black}image matching strategies} namely: the FLANN matcher \citep{muja2009fast} and brute-force matcher. 
As shown in Fig.~\ref{fig:eval_matching}, the brute-force matcher's timing is significantly lower than FLANN.
For KITTI 00, the average score is 3.32 $ms$, while for FLANN is 40.70 $ms$.
Respectively, for New College, the timings are 0.91 $ms$ and 15.62 $ms$.
This happens due to the proposed local features' low dimension {\color{black}(40-dimensional)}.

\subsection{Evaluating Deep Global Features}
\label{sec_gf}

A comparison of our global feature against two other contemporary CNN-based features is presented.
NetVLAD \citep{arandjelovic2016netvlad} and Resnet50-AP-GeM \citep{revaud2019learning} have been selected since these are the features commonly used as feature extractors in place recognition.
For our experiments, features extracted from NetVLAD and Resnet50-AP-GeM replaced our global representations.
However, the other modules remain the same.
As we can see in Table~\ref{table_gf}, the features provided by FILD++ achieve the highest recall rate in most of the evaluated datasets. 
We also recorded the timing needed for feature extraction when different extractors are used in Table~\ref{table_timeglobal}.
Our method requires only 11.23 $ms$ when applied on City Center, while NetVLAD and Resnet50-AP-GeM need 105.60 $ms$ and  22.60 $ms$, respectively.
The results show that our global feature outperforms the other methods in terms of speed and recall.

\begin{table}[!b]
\centering
\caption{Recalls at 100\% Precision: A Comparison of Our Method with Different CNN-based Global Features}
\label{table_gf}
\resizebox{\columnwidth}{!}{
\begin{tabular}{p{1.3cm}p{1.5cm}p{1.3cm}p{1.3cm}p{1.3cm}}
\toprule
\textbf{Dataset} &   &  \textbf{NetVLAD \citep{arandjelovic2016netvlad}} & \textbf{Resnet50-AP-GeM \citep{revaud2019learning}} & \textbf{FILD++}  \\
\toprule
KITTI & Seq\# 00             &   91.88                   &    91.24         &      \textbf{94.92}           \\
  & Seq\# 02             &   \textbf{74.77}                  &      73.21      &   73.52       \\
  & Seq\# 05             &   91.81                     &     94.70      &     \textbf{95.42}      \\
  & Seq\# 06             &    \textbf{98.90}                   &     97.79      &      98.16      \\
\hline
Oxford & New College             &   83.35                     &    \textbf{84.85}        &   82.37            \\
  & City Center              &      89.84                  &       \textbf{90.56}     &     90.01     \\
\hline
Malaga 2009& Parking 6L                       &  59.83                        &     60.11         &     \textbf{62.74}     \\
\hline
St. Lucia & 100909 (12:10)             &    80.46                    &    79.26       &      \textbf{83.39}        \\
 & 100909 (14:10)              &        63.80                  &       58.10     &    \textbf{66.41}      \\
 & 180809 (15:45)              &     79.67                   &    69.36         &    \textbf{81.36}    \\
 & 190809 (08:45)            &       83.21                 &   82.91         &      \textbf{87.86}  \\
\toprule
\end{tabular}
}
\end{table}

\begin{table}[!b]
\caption{\color{black}Average Feature Extraction Time (ms) Comparison of Our Method with Different CNN-based Global Features}
\label{table_timeglobal}
\setlength{\abovecaptionskip}{2pt}
\begin{center}
\resizebox{\columnwidth}{!}{
\begin{tabular}{c|p{1cm}|p{1cm}|p{1.2cm}|p{1.2cm}}
\toprule
Methods  & KITTI 00  & City Center  & Malaga6L   & St. Lucia 100909 (1210) \\
\midrule
\textbf{NetVLAD \citep{arandjelovic2016netvlad}} & 105.60 & 94.25 & 131.07 & 85.97 \\
\hline
\textbf{Resnet50-AP-GeM \citep{revaud2019learning}} & 22.60 & 19.90 & 35.33 & 16.93 \\
\hline
\textbf{Proposed Global Feature} & \textbf{11.23} & \textbf{8.38} & \textbf{17.25} & \textbf{8.54} \\
\bottomrule
\end{tabular}
}
\end{center}
\end{table}

\subsection{FILD++ Performance }
\label{sec_performance}

\begin{figure}[t]
\centering
 \includegraphics[width=0.48\textwidth]{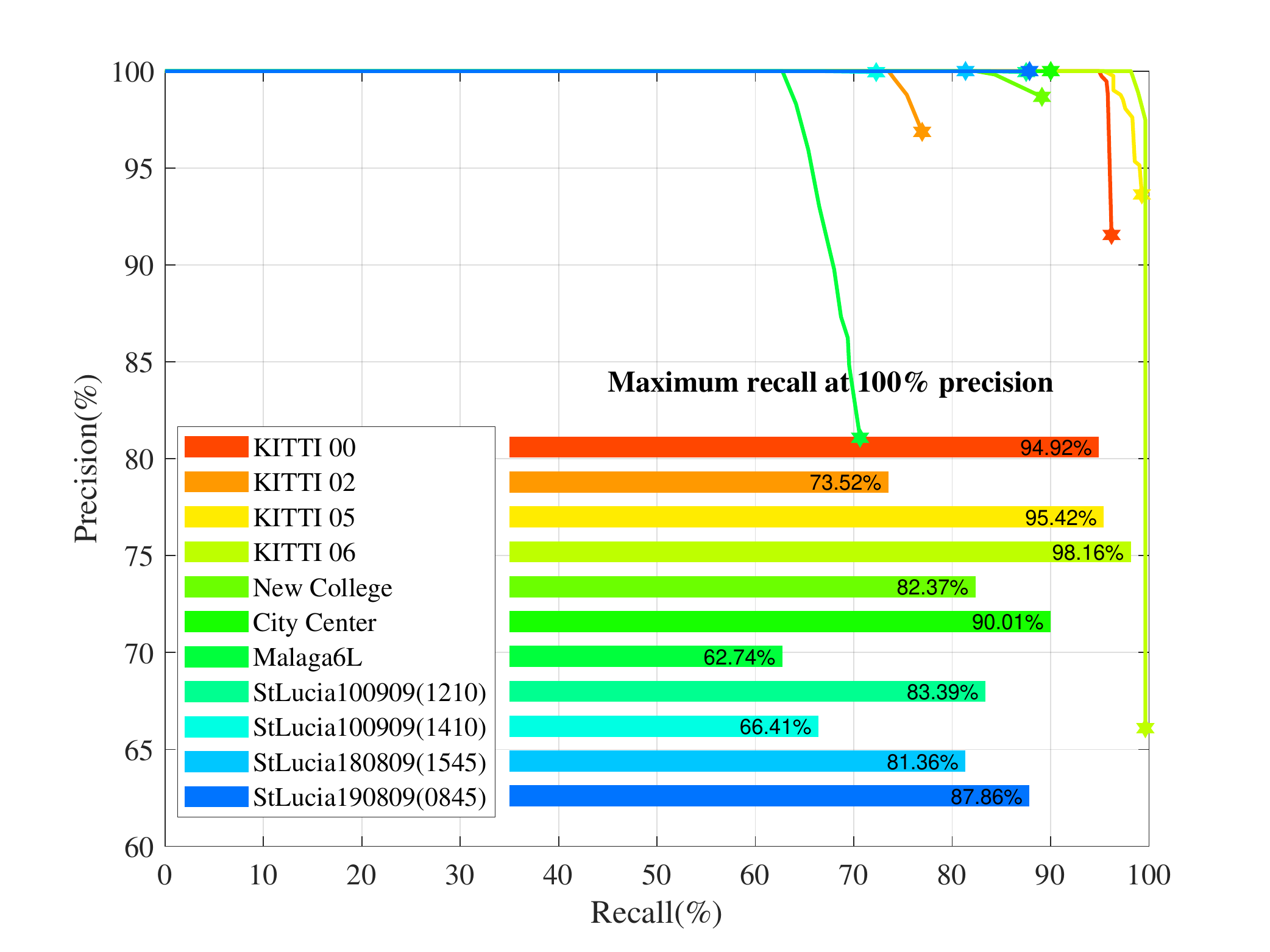}
  \caption{Our algorithm's precision-recall curves on each evaluated dataset.}
\label{fig:gt_perf}
\end{figure}

\begin{figure}[t]
\centering
 \includegraphics[width=0.48\textwidth]{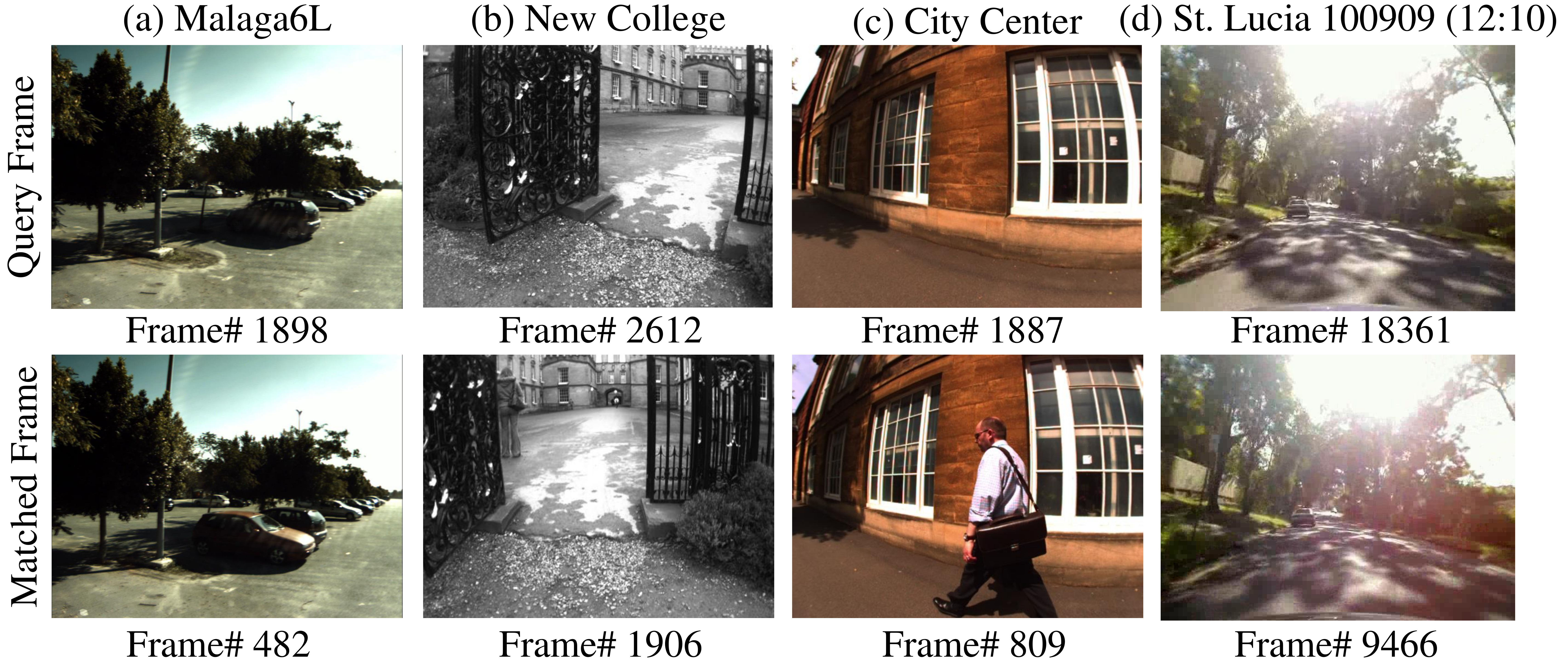}
  \caption{Some example images of the detected loop-closure locations. }
\label{fig:gt_hardcase}
\end{figure}
\begin{figure}[!t]
\centering
 \includegraphics[width=0.48\textwidth]{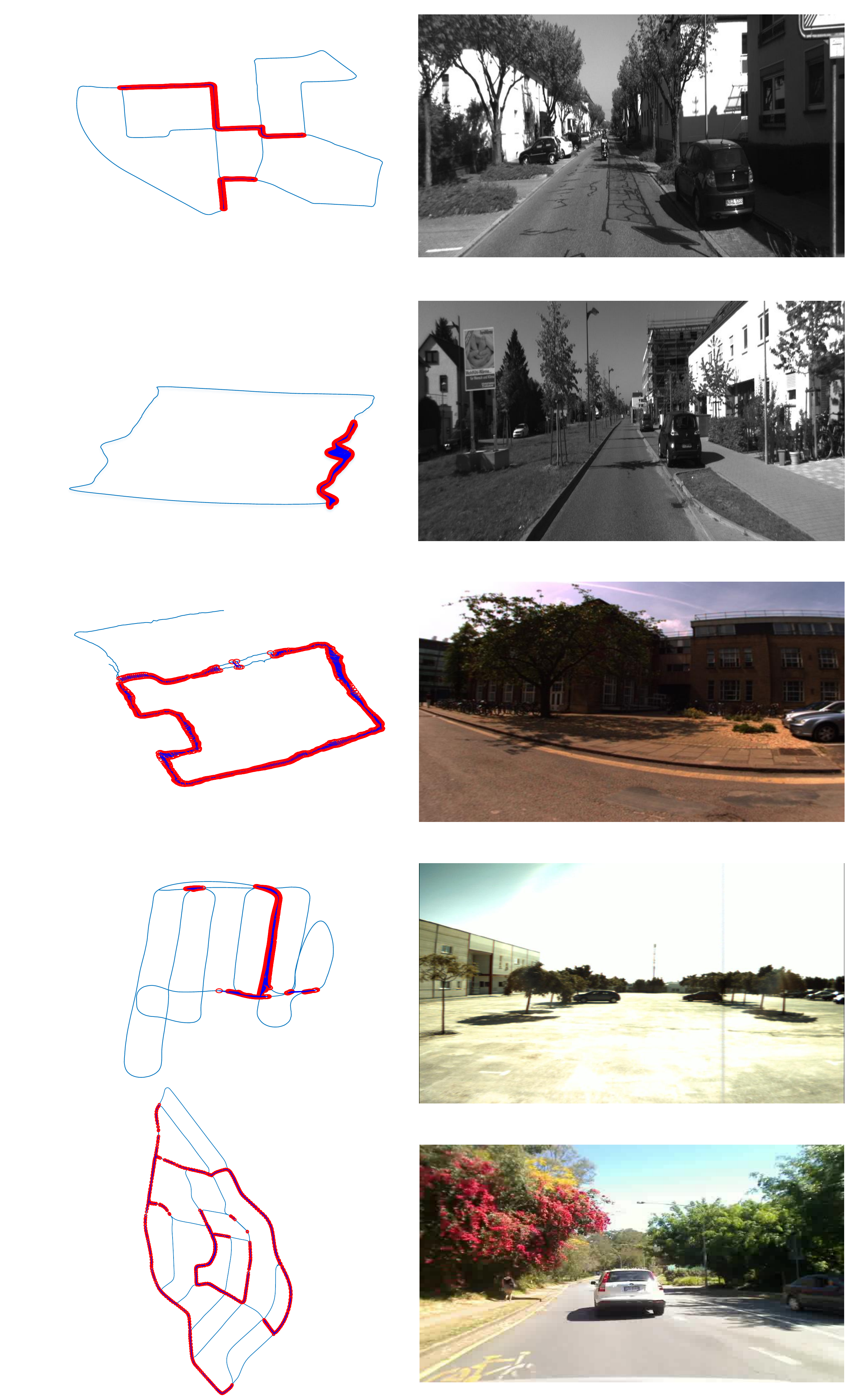}
  \caption{Robot trajectories (left) and example images (right). From top to bottom: KITTI 00 \citep{Geiger2012CVPR}, KITTI 06 \citep{Geiger2012CVPR}, City Center \citep{cummins2008fab}, Malaga6L \citep{blanco2009collection}, St. Lucia 100909 (12:10) \citep{glover2010fab}. The loop closure detections are labelled using red circles. }
\label{fig:path}
\end{figure}

\color{black}
In Table~\ref{table_all}, we list our system's highest recall score at 100\% precision on eleven datasets, while compared to the baseline methods.
As shown FILD++ outperforms the other methods on eight out of eleven image-sequences.
Malaga6L is recorded at a parking site, thus presents high scene similarity due to the absence of distinct differences between the roads.
Therefore, each of the evaluated methods performs poorly.
As far as the KITTI vision suite and Oxford datasets are concerned, improved performance is demonstrated.
This is mostly owing to architectural constructions appearing in these environments, which are similar to the training set of our feature extraction network.
Hence, our pipeline can extract more representative deep features and compare them more precisely for these types of scenes.
\color{black}

Fig.~\ref{fig:gt_perf} illustrates the precision-recall curves generated by varying the number of RANSAC inliers.
Our framework can successfully detect loops through a recall score ranging from 62.74\% (Malaga6L) to 98.16\% (KITTI 06).
Malaga6L is the most challenging dataset and KITTI 06 is the smallest dataset among the rest.
Some examples of TP detections are shown in Fig.~\ref{fig:gt_hardcase}.
It is worth noting that when dynamic objects are included, \textit{e.g.,} cars in Fig.~\ref{fig:gt_hardcase}~(a) and people in Fig.~\ref{fig:gt_hardcase} (c), FILD++ can correctly identify the pre-visited location.
The example in Fig.~\ref{fig:gt_hardcase}~(b) demonstrate that our system can handle the viewpoint changes, while Fig.~\ref{fig:gt_hardcase}~(d) shows its ability to deal with illumination variations.
\color{black}
We show the loop closure detections detected by our framework on to of the robot’s trajectories in Fig.~\ref{fig:path}.
\color{black}

\subsection{Time Requirements}
\label{sec_time}

\begin{table*}[t]
\centering
\caption{Recalls at 100\% Precision: A Comparison of The Baseline Methods with Our Framework}
\label{table_all}
\resizebox{\textwidth}{!}{
\begin{tabular}{p{1.8cm}p{2cm}p{2.0cm}p{1.4cm}p{1.4cm}p{1.5cm}p{1.5cm}ll}
\toprule
\textbf{Dataset} &   &  \textbf{DLoopDetector \citep{galvez2012bags}$^{\dagger}$} & \textbf{Tsintotas et al. \citep{tsintotas2018assigning}} & \textbf{PREVIeW \citep{bampis2018fast}$^{\ddagger}$}  & \textbf{iBoW-LCD \citep{garcia2018ibow}$^{\mathsection}$} & \textbf{Kazmi et al. \citep{kazmi2019detecting}}$^{\mathparagraph}$ & \textbf{FILD \citep{anshan2019}}  & \textbf{FILD++} \\
\toprule
KITTI & Seq\# 00             &   72.43                   &    93.18         &      89.47       &    76.50    &  90.39  &   91.23  &    \bfseries94.92     \\
  & Seq\# 02             &   68.22                  &      76.01      &   71.96         &    72.22    &    \bfseries79.49    &    65.11   &    73.52  \\
  & Seq\# 05             &   51.97                     &     94.20      &     87.71        &   53.07    &    81.41    &  85.15    &    \bfseries95.42    \\
  & Seq\# 06             &    89.71                   &     86.03      &      80.15        &    95.53     &  97.39   &    93.38    &  \bfseries98.16   \\
\hline
Oxford & New College             &   47.56                     &    52.44        &   80.87             & 73.14    &  51.09 &   76.74          & \bfseries82.37   \\
  & City Center              &      30.59                  &       16.30     &     49.63       &    82.03      &    75.58  &    66.48    &   \bfseries90.01   \\
\hline
Malaga 2009& Parking 6L                       &  31.02                        &     59.14         &     33.93       &  57.48  &  50.98    &  56.09 & \bfseries62.74 \\
\hline
St. Lucia & 100909 (12:10)             &    37.22                     &    26.27       &      60.93           &    70.02   &   80.06  &    76.06    &   \bfseries83.39  \\
 & 100909 (14:10)              &        14.87                  &       9.77     &    23.06       &      \bfseries68.06     &  58.10    &     53.84   &   66.41  \\
 & 180809 (15:45)              &     31.36                   &    15.07         &    49.79       &      \bfseries87.50   &    72.55    &     66.96    &   81.36  \\
 & 190809 (08:45)            &       39.78                 &    27.68         &      56.69      &       59.36   &  80.13   &     78.00      &  \bfseries87.86 \\
\toprule
\end{tabular}
}
\flushleft{}
\footnotesize{$^{\dagger}$ Compared to \citep{galvez2012bags}, we use different number of images for New College and Malaga6L.
We have changed the normalized similarity score threshold to achieve 100\% precision, as there are false detections using the default parameters.
$^{\ddagger}$ We report the recall using the default parameters. However, the precision of each dataset cannot achieve 100\%.
$^{\mathsection}$ We report the iBoW-LCD recalls on KITTI dataset from \citep{kazmi2019detecting}.
$^{\mathparagraph}$ We quote the results as reported in~\citep{kazmi2019detecting}, as an open-source implementation were not available.
 } \\
\end{table*}

\begin{table}[t]
	\caption{Average Execution Time (ms/query) on The Representative Datasets}
	\label{table_methodstimecost}
	\setlength{\abovecaptionskip}{2pt}
	\begin{center}
\resizebox{\columnwidth}{!}{
	\begin{tabular}{c|p{1cm}|p{1cm}|p{1.2cm}|p{1.2cm}}
	\toprule
	Approach  & KITTI 00  & City Center  & Malaga6L   & St. Lucia 100909 (1210) \\
	\midrule
	DLoopDetector \citep{galvez2012bags} & 111.04 & 27.51 & 42.57 & 91.04 \\
	\hline
Tsintotas et al. \citep{tsintotas2018assigning} & 521.54 & 183.23 & 638.61 & 625.05 \\
	\hline
	PREVIeW \citep{bampis2018fast} & 32.39 & 34.09 & 36.33 & 25.40 \\
	\hline
	FILD \citep{anshan2019} & 62.68 & 40.23 & 68.16 & 49.10 \\
	\hline
	FILD++ &  38.70 & 32.10 & 56.56 & 34.20 \\
	\bottomrule
	\end{tabular}
}
	\end{center}
\end{table}

\begin{table}[t]
\caption{Average Execution Time (ms/query) of Our method in Different Datasets}
	\label{table_timecost}
	\setlength{\abovecaptionskip}{2pt}
	\begin{center}
\resizebox{\columnwidth}{!}{
	\begin{tabular}{c|p{1cm}|p{1cm}|p{1.2cm}|p{1.2cm}}
	\toprule
	Stages  & KITTI 00  & City Center  & Malaga6L   & St. Lucia 100909 (1210) \\
	\midrule
	Feature Extraction & 29.67 & 22.04 & 45.41 & 22.48 \\
	\hline
	Adding Feature & 4.69 & 4.30 & 3.63 & 6.03 \\
	\hline
	Graph Searching & 0.64 & 0.33 & 0.47 & 0.70 \\
	\hline
	Feature Matching & 3.32& 2.17 & 4.13 & 2.26  \\
	\hline
	RANSAC & 0.38 & 3.26 & 2.92 & 2.73 \\
	\hline
	Whole System & 38.70 & 32.10 & 56.56 & 34.20 \\
	\bottomrule
	\end{tabular}
}
	\end{center}
\end{table}

\begin{figure}[t]
\centering
 \includegraphics[width=0.48\textwidth]{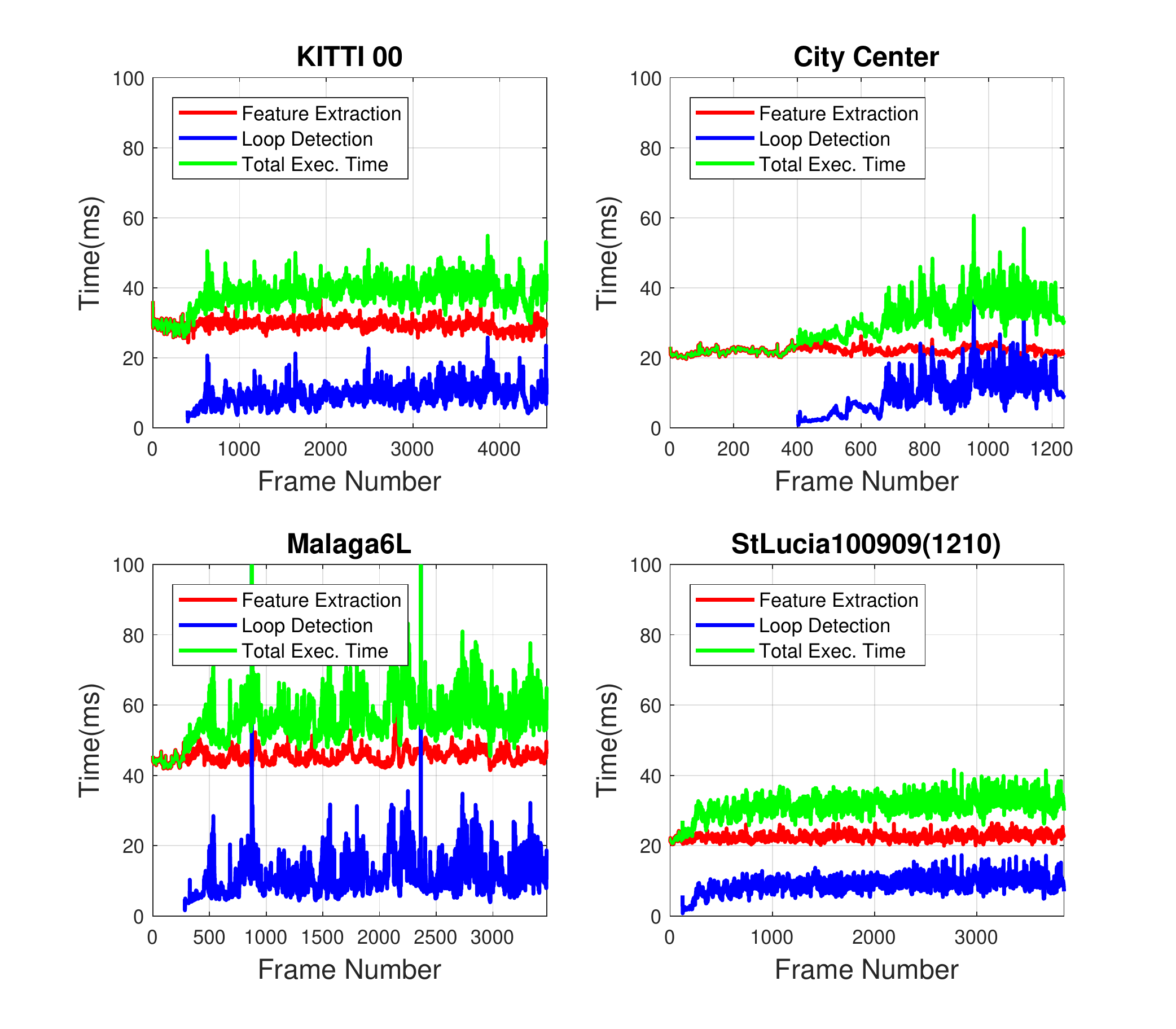}
  \caption{Execution times of our algorithm.
  }
\label{fig:eval_time}
\end{figure}

\begin{table}[t]
	\caption{Average Execution Time (ms/query) in New College with 52480 Images}
	\label{table_timecost52480}
	\setlength{\abovecaptionskip}{2pt}
	\begin{center}
\resizebox{\columnwidth}{!}{
	\begin{tabular}{c|p{1cm}|p{1cm}|p{1cm}|p{1.2cm}}
	\toprule
Stages  & Mean  & Std  & Max   & Min \\
	\midrule
	Feature Extraction & 14.62 & 0.65 & 21.13 & 12.30 \\
	\hline
	Adding Feature & 3.97 & 2.47 & 22.63 & 0.04  \\
	\hline
	Graph Searching & 0.67 & 0.19 & 3.20 & 0.04 \\
	\hline
	Feature Matching & 1.06 & 0.10 & 14.04 & 0.08 \\
	\hline
	RANSAC & 1.72 & 1.08 & 17.01 & 0.0 \\
	\hline
	Whole System & 22.05 & 5.04 & 58.98 & 14.56  \\
	\bottomrule
	\end{tabular}
}
	\end{center}
\end{table}

\color{black}
We have estimated our system's complexity on four representative datasets.
As shown in Table~\ref{table_methodstimecost}, FILD++ achieves a higher speed than its predecessor.
In general, this improvement is owing to the local features' low dimensionality which permits faster image matching.
\color{black}

\begin{figure}[t]
\centering
 \includegraphics[width=0.5\textwidth]{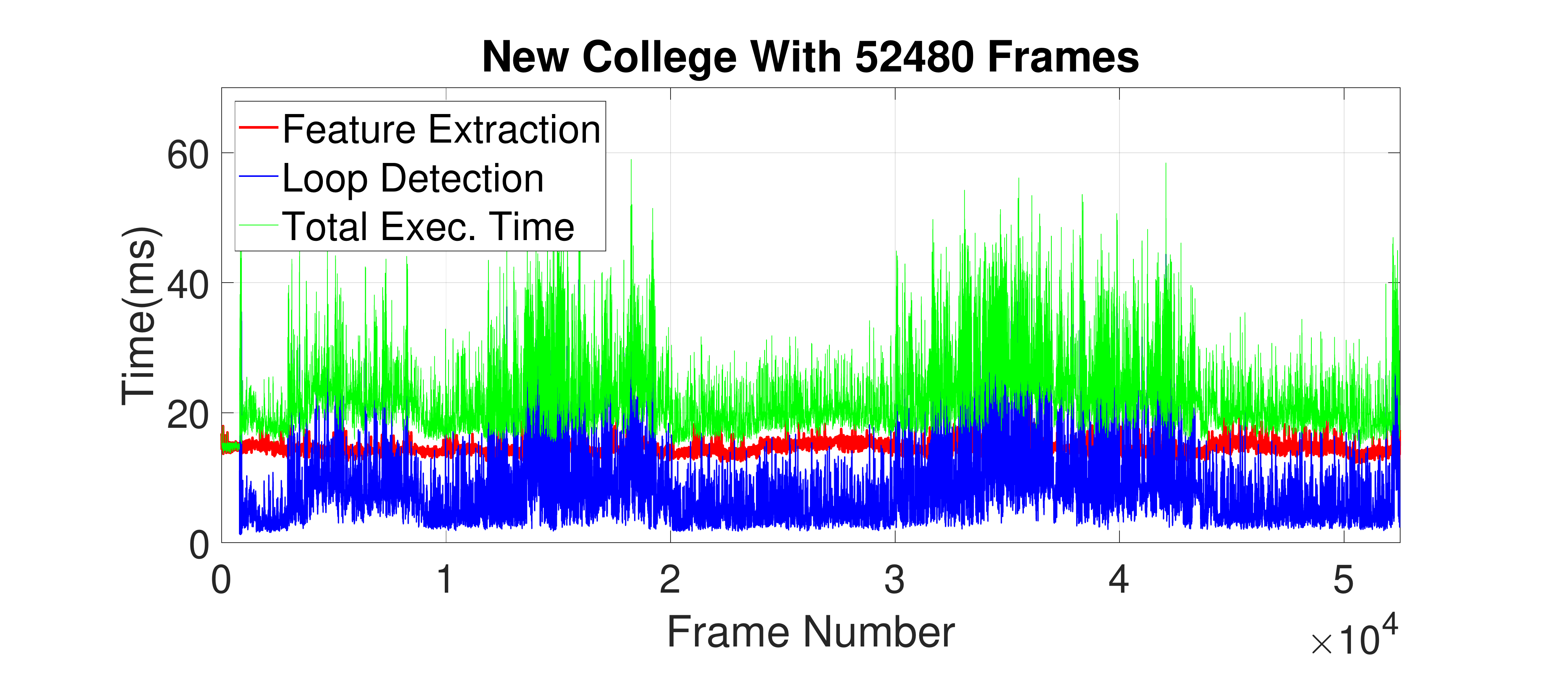}
  \caption{Execution times in New College with 52480 images.}
\label{fig:eval_newcollegetime}
\end{figure}

The average execution times for different pipeline stages are presented in Table~\ref{table_timecost}.
Also, in Fig.~\ref{fig:eval_time}, we present the timings for features' extraction and loops' detection as a function of  frame number.
As illustrated, FILD++ requires constant time for each dataset, while
the features' extraction is the most costly procedure.
For Malaga6L, our pipeline needs about 50 $ms$ to 80 $ms$ for the total execution time, while the feature extraction requires about 45 $ms$.
This happens due to the images' resolution, which is the largest among the evaluated datasets.
Concurrently, for St. Lucia, the average timing is below 40 $ms$, because of the different image resolution.
Furthermore, it is observed that the timing for our indexing graph-based technique is below 1 $ms$ and the whole system's speed ranges from 32 $ms$ to 57 $ms$ demonstrating FILD++'s high efficiency.
In Table~\ref{table_timecost52480}, we test our system's scalability setting the frequency of New College to $f=20$ $Hz$ and obtained 52480 images.
The average execution time is about 22 $ms$.
As can be seen in Fig.~\ref{fig:eval_newcollegetime}, an increase of frames number would not induce a rise of processing time.

\begin{figure}[t]
\centering
 \includegraphics[width=0.5\textwidth]{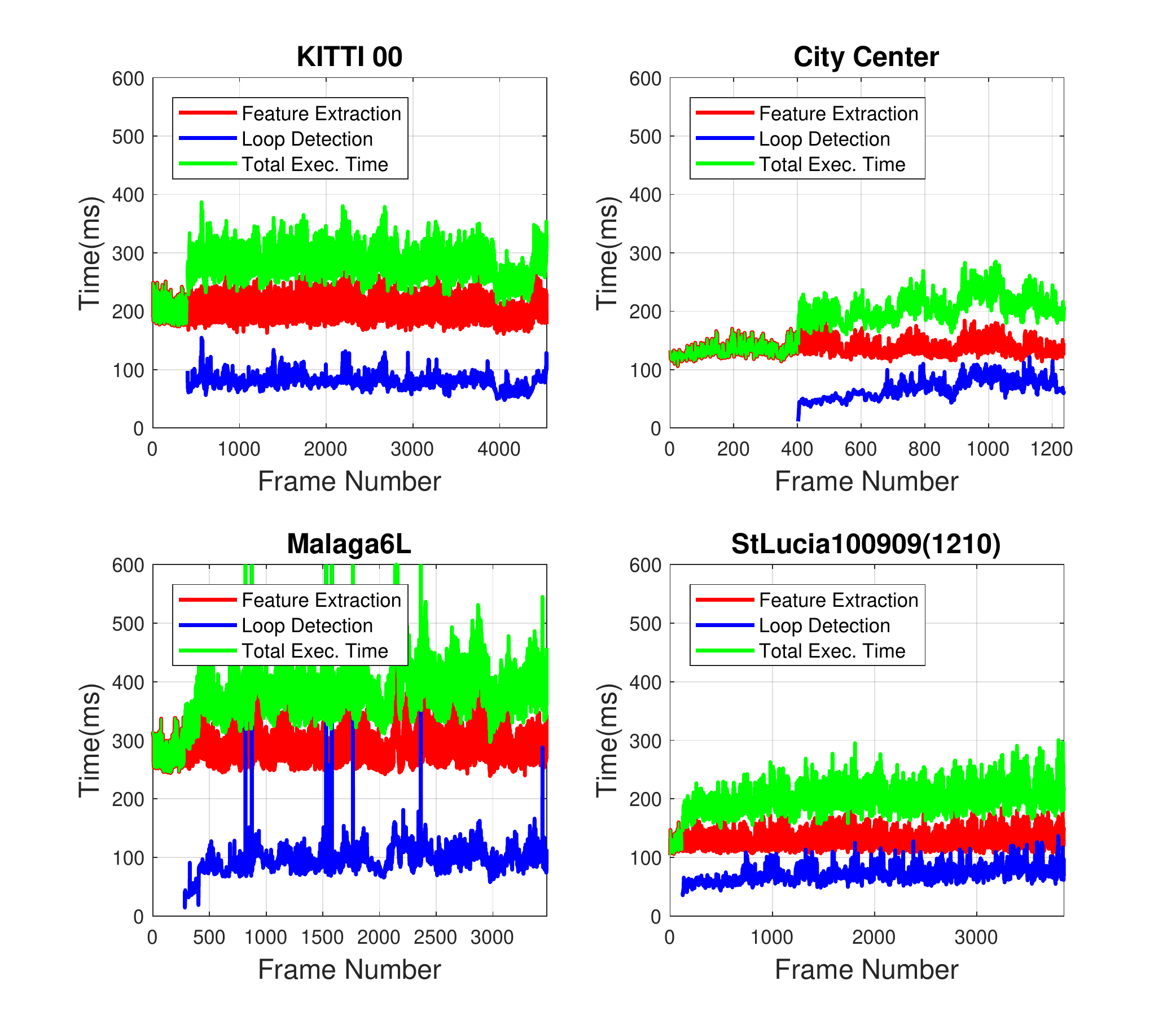}
  \caption{Execution times on an NVIDIA Jetson TX2 GPU.}
\label{fig:eval_tx2time}
\end{figure}

\begin{table}[t]
\caption{Average Execution Time (ms/query) on an NVIDIA Jetson TX2 GPU}
\label{table_tx2timecost}
\setlength{\abovecaptionskip}{2pt}
\begin{center}
\resizebox{\columnwidth}{!}{
\begin{tabular}{c|p{1cm}|p{1cm}|p{1.2cm}|p{1.2cm}}
\toprule
Stages  & KITTI 00  & City Center  & Malaga6L   & St. Lucia 100909 (1210) \\
\midrule
Feature Extraction & 200.12 & 135.97 & 292.06 & 128.19 \\
\hline
Loop Detection & 78.84 & 68.44 & 96.60 & 72.39 \\
\hline
Whole System & 278.96 & 204.41 & 388.66 & 200.58 \\
\bottomrule
\end{tabular}
}
\end{center}
\end{table}

We also implemented our algorithm on the Jetson TX2 platform in Max-N mode (all CPU cores in use and GPU clocked at $1.3$ $GHz$) and show the timing in Fig.~\ref{fig:eval_tx2time}.
FILD++ does not require extra processing time even if applied in an embedded platform.
The most time-consuming stage is the features' extraction as we perform two forward passes for each image frame.
In Table~\ref{table_tx2timecost}, we list the average time for the feature extraction, the loop detection and the whole system.
The proposed system processes Malaga6L in 388.66 $ms$, while for any other dataset, processing times are below 300 $ms$ indicating its low computational complexity.

\section{Discussion and Conclusion}
\label{conclusion}

In this article, a visual loop closure detection approach is proposed, dubbed as FILD++.
Through two forward passes of a single network, our system extracts global and local deep features for filtering and re-ranking, respectively.
Along with the robot's navigation, an HNSW graph is built incrementally based on the global features permitting fast indexing and database search during query.
When a candidate location is retrieved it is geometrically verified using the provided local features.
Eleven publicly-available datasets are chosen for our evaluation showing FILD++'s effectiveness and efficiency compared with other state-of-the-art approaches.

\color{black}
The proposed FILD++ framework has three advantages compared with the previous FILD method. 
Firstly, the proposed framework is more compact. 
This is because only one network was used for feature extraction. 
In addition, the extracted deep local features are only 40-dimensional, which is significantly lower than SURF (128-dimensional). 
Because there is only one network and without the usage of CasHash~\citep{cheng2014fast}, the source code of FILD++ is more concise than FILD, as given in the GitHub\footnote{https://github.com/anshan-ar/FILD}. 
Besides, the dimension of global feature in FILD++ is also lower than that in FILD, which is 1024-dimensional \textit{vs.} 1280-dimensional. 

Secondly, the proposed method is simpler than the previous method. 
For feature extraction, FILD extracts global features using MobileNetV2 and local features using SURF. 
We simplify the feature extraction process in this work. 
The deep global features and local features are extracted via two forward passes of a single network. 
This dramatically simplifies the feature extraction process. 
Because the dimension of the deep local feature extracted by our method is only 40-dimensional, we can use a brute-force matcher for efficient feature matching. 
Therefore we did not use CasHash~\citep{cheng2014fast} in FILD++.
As a result, the hash code creation process is unnecessary, which simplifies the whole process.

Last but not least, FILD++ is much faster than its previous version. 
As shown in Table~\ref{table_methodstimecost}, FILD++ costs 38.70 $ms$ per query on KITTI 00 dataset, while FILD requires 62.68 $ms$. 
Thus, it can be seen that FILD++ is significantly faster than FILD on all datasets. 
Table~\ref{table_timecost52480_all} also shows the average execution time of FILD and FILD++ in the New College dataset (52480 Images). 
As can be seen, the feature extraction in FILD needs more time than in FILD++. 
The hash codes creation step in FILD is also time-consuming, while there is no such step in FILD++. 
Because SURF features in FILD are different from the deep local features, we extracted in FILD++, the timing for RANSAC scheme is different. 
We can see our approach also takes less time at this step. 
The overall time cost of the proposed FILD++ is 22.05 $ms$ per query, while for FILD is 50.28 $ms$. 
This indicates the speed advantage of our new method when applied in large datasets.   

\begin{table}[t]
	\caption{\textcolor{black}{Average Execution Time (ms/query) of FILD \citep{anshan2019} and FILD++ in New College with 52480 Images}}
	\label{table_timecost52480_all}
	\setlength{\abovecaptionskip}{2pt}
	\begin{center}
\resizebox{\columnwidth}{!}{
	\begin{tabular}{c|c|c}
	\toprule
Method  & FILD~\citep{anshan2019}  & FILD++ \\
	\midrule
	Feature Extraction & 17.69 & 14.62  \\
	\hline
	Hash Codes Creation & 16.94 & 0.0   \\
	\hline
	Adding Feature & 5.21 & 3.97   \\
	\hline
	Graph Searching & 0.93 & 0.67 \\
	\hline
	Feature Matching & 2.23 & 1.06  \\
	\hline
	RANSAC & 7.55 & 1.72  \\
	\hline
	Whole System & 50.28 & 22.05 \\
	\bottomrule
	\end{tabular}
}
	\end{center}
\end{table}

\color{black}

Our system's performance depends on several factors: the reliability of its deep features, the HNSW's retrieval precision, and the effectiveness of the geometrical verification.
The similarity scores of the query and the candidate images are not utilized.
A proper threshold may have helped us with FP elimination; however, the complexity of the system would be inevitably high.
During geometrical verification, as the number of matches $n$ is an important parameter, the easiest way to achieve a higher recall is to increase its value.
Howbeit, as illustrated, such action is time-demanding; therefore, a convenient trade-off is considered.

Our plans include the integration of the proposed method to a SLAM framework, while an increase of the classification accuracy will lead to higher performance.
Consequently, using more powerful networks, such as ResNeXt \citep{xie2017aggregated} and ResNeSt \citep{zhang2020resnest}, we should be able to improve the system's performance.

\section*{Acknowledgment}

The authors wish  to gratefully acknowledge Dr. Mark Cummins for his kindly help and Guangfu Che, whose constructive suggestions helped the system evaluation.

This work was funded by grants from the National Key Research and Development Program of China (Grant No. 2020YFC2006200).

\printendnotes

\bibliography{bible}

\begin{thebibliography}{89}
\expandafter\ifx\csname natexlab\endcsname\relax\def\natexlab#1{#1}\fi
\expandafter\ifx\csname url\endcsname\relax
  \def\url#1{\texttt{#1}}\fi
\expandafter\ifx\csname urlprefix\endcsname\relax\def\urlprefix{URL: }\fi

\bibitem[{Amanatiadis et~al.(2011)Amanatiadis, Kaburlasos, Gasteratos and
  Papadakis}]{amanatiadis2011evaluation}
Amanatiadis, A., Kaburlasos, V., Gasteratos, A. and Papadakis, S. (2011)
  Evaluation of shape descriptors for shape-based image retrieval.
\newblock \textit{IET Image Process.}, \textbf{5}, 493--499.

\bibitem[{An et~al.(2019)An, Che, Zhou, Liu, Ma and Chen}]{anshan2019}
An, S., Che, G., Zhou, F., Liu, X.~L., Ma, X. and Chen, Y. (2019) Fast and
  incremental loop closure detection using proximity graphs.
\newblock In \textit{IEEE/RSJ Int. Conf. Intell. Robots Syst.}, 378--385.

\bibitem[{Andoni and Razenshteyn(2015)}]{andoni2015optimal}
Andoni, A. and Razenshteyn, I. (2015) Optimal data-dependent hashing for
  approximate near neighbors.
\newblock In \textit{Proc. ACM Symp. Theory of computing}, 793--801.

\bibitem[{{Angeli} et~al.(2008){Angeli}, {Filliat}, {Doncieux} and
  {Meyer}}]{angeli2008fast}
{Angeli}, A., {Filliat}, D., {Doncieux}, S. and {Meyer}, J. (2008) Fast and
  incremental method for loop-closure detection using bags of visual words.
\newblock \textit{IEEE Trans. Robot.}, \textbf{24}, 1027--1037.

\bibitem[{Arandjelovic et~al.(2016)Arandjelovic, Gronat, Torii, Pajdla and
  Sivic}]{arandjelovic2016netvlad}
Arandjelovic, R., Gronat, P., Torii, A., Pajdla, T. and Sivic, J. (2016)
  Netvlad: Cnn architecture for weakly supervised place recognition.
\newblock In \textit{IEEE Conf. Comp. Vis. Pattern Recogn.}, 5297--5307.

\bibitem[{Babenko et~al.(2014)Babenko, Slesarev, Chigorin and
  Lempitsky}]{babenko2014neural}
Babenko, A., Slesarev, A., Chigorin, A. and Lempitsky, V. (2014) Neural codes
  for image retrieval.
\newblock In \textit{Euro. Conf. Comp. Vis.}, 584--599. Springer.

\bibitem[{Bampis et~al.(2016)Bampis, Amanatiadis and
  Gasteratos}]{bampis2016encoding}
Bampis, L., Amanatiadis, A. and Gasteratos, A. (2016) {Encoding the description
  of image sequences: A two-layered pipeline for loop closure detection}.
\newblock In \textit{IEEE/RSJ Int. Conf. Intell. Robots Syst.}, 4530--4536.

\bibitem[{Bampis et~al.(2018)Bampis, Amanatiadis and
  Gasteratos}]{bampis2018fast}
--- (2018) Fast loop-closure detection using visual-word-vectors from image
  sequences.
\newblock \textit{Int. J. Robot. Res.}, \textbf{37}, 62--82.

\bibitem[{Bay et~al.(2006)Bay, Tuytelaars and Van~Gool}]{bay2006surf}
Bay, H., Tuytelaars, T. and Van~Gool, L. (2006) Surf: Speeded up robust
  features.
\newblock In \textit{Euro. Conf. Comp. Vis.}, 404--417.

\bibitem[{Blanco et~al.(2009)Blanco, Moreno and
  Gonzalez}]{blanco2009collection}
Blanco, J.-L., Moreno, F.-A. and Gonzalez, J. (2009) A collection of outdoor
  robotic datasets with centimeter-accuracy ground truth.
\newblock \textit{Autonomous Robots}, \textbf{27}, 327.

\bibitem[{Bosch et~al.(2007)Bosch, Zisserman and Munoz}]{bosch2007representing}
Bosch, A., Zisserman, A. and Munoz, X. (2007) Representing shape with a spatial
  pyramid kernel.
\newblock In \textit{Proc. ACM Int. Conf. Image and Video Retrieval}, 401--408.
  ACM.

\bibitem[{Botterill et~al.(2011)Botterill, Mills and Green}]{botterill2011bag}
Botterill, T., Mills, S. and Green, R. (2011) Bag-of-words-driven,
  single-camera simultaneous localization and mapping.
\newblock \textit{Journal of Field Robotics}, \textbf{28}, 204--226.

\bibitem[{Cadena et~al.(2016)Cadena, Carlone, Carrillo, Latif, Scaramuzza,
  Neira, Reid and Leonard}]{cadena2016past}
Cadena, C., Carlone, L., Carrillo, H., Latif, Y., Scaramuzza, D., Neira, J.,
  Reid, I. and Leonard, J.~J. (2016) Past, present, and future of simultaneous
  localization and mapping: Toward the robust-perception age.
\newblock \textit{IEEE Trans. Robot.}, \textbf{32}, 1309--1332.

\bibitem[{Calonder et~al.(2010)Calonder, Lepetit, Strecha and
  Fua}]{calonder2010brief}
Calonder, M., Lepetit, V., Strecha, C. and Fua, P. (2010) Brief: Binary robust
  independent elementary features.
\newblock In \textit{Euro. Conf. Comp. Vis.}, 778--792.

\bibitem[{Cascianelli et~al.(2017)Cascianelli, Costante, Bellocchio, Valigi,
  Fravolini and Ciarfuglia}]{cascianelli2017robust}
Cascianelli, S., Costante, G., Bellocchio, E., Valigi, P., Fravolini, M.~L. and
  Ciarfuglia, T.~A. (2017) Robust visual semi-semantic loop closure detection
  by a covisibility graph and cnn features.
\newblock \textit{Robotics and Autonomous Systems}, \textbf{92}, 53--65.

\bibitem[{Chan et~al.(2015)Chan, Jia, Gao, Lu, Zeng and Ma}]{chan2015pcanet}
Chan, T.-H., Jia, K., Gao, S., Lu, J., Zeng, Z. and Ma, Y. (2015) {PCANet}: A
  simple deep learning baseline for image classification?
\newblock \textit{IEEE Trans. Image Process.}, \textbf{24}, 5017--5032.

\bibitem[{Chanc{\'a}n et~al.(2020)Chanc{\'a}n, Hernandez-Nunez, Narendra,
  Barron and Milford}]{chancan2020hybrid}
Chanc{\'a}n, M., Hernandez-Nunez, L., Narendra, A., Barron, A.~B. and Milford,
  M. (2020) A hybrid compact neural architecture for visual place recognition.
\newblock \textit{IEEE Robot. Autom. Lett.}, \textbf{5}, 993--1000.

\bibitem[{Chen et~al.(2017)Chen, Jacobson, S{\"u}nderhauf, Upcroft, Liu, Shen,
  Reid and Milford}]{chen2017deep}
Chen, Z., Jacobson, A., S{\"u}nderhauf, N., Upcroft, B., Liu, L., Shen, C.,
  Reid, I. and Milford, M. (2017) Deep learning features at scale for visual
  place recognition.
\newblock In \textit{IEEE Int. Conf. Robot. Autom.}, 3223--3230.

\bibitem[{Chen et~al.(2018)Chen, Liu, Sa, Ge and Chli}]{chen2018learning}
Chen, Z., Liu, L., Sa, I., Ge, Z. and Chli, M. (2018) Learning context flexible
  attention model for long-term visual place recognition.
\newblock \textit{IEEE Robot. Autom. Lett.}, \textbf{3}, 4015--4022.

\bibitem[{Cheng et~al.(2014)Cheng, Leng, Wu, Cui and Lu}]{cheng2014fast}
Cheng, J., Leng, C., Wu, J., Cui, H. and Lu, H. (2014) Fast and accurate image
  matching with cascade hashing for 3d reconstruction.
\newblock In \textit{IEEE Conf. Comp. Vis. Pattern Recogn.}, 1--8.

\bibitem[{Chow and Liu(1968)}]{chow1968approximating}
Chow, C. and Liu, C. (1968) Approximating discrete probability distributions
  with dependence trees.
\newblock \textit{IEEE Trans. Infor. Theory}, \textbf{14}, 462--467.

\bibitem[{Cummins and Newman(2008)}]{cummins2008fab}
Cummins, M. and Newman, P. (2008) {FAB-MAP: Probabilistic localization and
  mapping in the space of appearance}.
\newblock \textit{Int. J. Robot. Res.}, \textbf{27}, 647--665.

\bibitem[{Cummins and Newman(2011)}]{cummins2011appearance3}
--- (2011) {Appearance-only SLAM at large scale with FAB-MAP 2.0}.
\newblock \textit{Int. J. Robot. Res.}, \textbf{30}, 1100--1123.

\bibitem[{Dugas et~al.(2001)Dugas, Bengio, B{\'e}lisle, Nadeau and
  Garcia}]{dugas2001incorporating}
Dugas, C., Bengio, Y., B{\'e}lisle, F., Nadeau, C. and Garcia, R. (2001)
  Incorporating second-order functional knowledge for better option pricing.
\newblock In \textit{Adv. Neural Inf. Process. Syst.}, 472--478.

\bibitem[{Engel et~al.(2015)Engel, St{\"u}ckler and Cremers}]{engel2015large}
Engel, J., St{\"u}ckler, J. and Cremers, D. (2015) Large-scale direct slam with
  stereo cameras.
\newblock In \textit{IEEE/RSJ Int. Conf. Intell. Robots Syst.}, 1935--1942.

\bibitem[{Filliat(2007)}]{filliat2007visual}
Filliat, D. (2007) A visual bag of words method for interactive qualitative
  localization and mapping.
\newblock In \textit{IEEE Int. Conf. Robot. Autom.}, 3921--3926.

\bibitem[{G{\'a}lvez-L{\'o}pez and Tard{\'o}s(2012)}]{galvez2012bags}
G{\'a}lvez-L{\'o}pez, D. and Tard{\'o}s, J.~D. (2012) Bags of binary words for
  fast place recognition in image sequences.
\newblock \textit{IEEE Trans. Robot.}, \textbf{28}, 1188--1197.

\bibitem[{Garcia-Fidalgo and Ortiz(2015)}]{garcia2015vision}
Garcia-Fidalgo, E. and Ortiz, A. (2015) Vision-based topological mapping and
  localization methods: A survey.
\newblock \textit{Robotics and Autonomous Systems}, \textbf{64}, 1--20.

\bibitem[{Garcia-Fidalgo and Ortiz(2017)}]{garcia2017hierarchical}
--- (2017) Hierarchical place recognition for topological mapping.
\newblock \textit{IEEE Trans. Robot.}, \textbf{33}, 1061--1074.

\bibitem[{Garcia-Fidalgo and Ortiz(2018)}]{garcia2018ibow}
--- (2018) {iBoW-LCD}: An appearance-based loop-closure detection approach
  using incremental bags of binary words.
\newblock \textit{IEEE Robot. Autom. Lett.}, \textbf{3}, 3051--3057.

\bibitem[{Gehrig et~al.(2017)Gehrig, Stumm, Hinzmann and
  Siegwart}]{gehrig2017visual}
Gehrig, M., Stumm, E., Hinzmann, T. and Siegwart, R. (2017) Visual place
  recognition with probabilistic voting.
\newblock In \textit{IEEE Int. Conf. Robot. Autom.}, 3192--3199.

\bibitem[{Geiger et~al.(2012)Geiger, Lenz and Urtasun}]{Geiger2012CVPR}
Geiger, A., Lenz, P. and Urtasun, R. (2012) Are we ready for autonomous
  driving? the kitti vision benchmark suite.
\newblock In \textit{IEEE Conf. Comp. Vis. Pattern Recogn.}, 3354--3361.

\bibitem[{Glover et~al.(2010)Glover, Maddern, Milford and
  Wyeth}]{glover2010fab}
Glover, A.~J., Maddern, W.~P., Milford, M.~J. and Wyeth, G.~F. (2010) Fab-map+
  ratslam: Appearance-based slam for multiple times of day.
\newblock In \textit{IEEE Int. Conf. Robot. Autom.}, 3507--3512. IEEE.

\bibitem[{Gordo et~al.(2016)Gordo, Almaz{\'a}n, Revaud and
  Larlus}]{gordo2016deep}
Gordo, A., Almaz{\'a}n, J., Revaud, J. and Larlus, D. (2016) Deep image
  retrieval: Learning global representations for image search.
\newblock In \textit{Euro. Conf. Comp. Vis.}, 241--257. Springer.

\bibitem[{Hajebi and Zhang(2014)}]{hajebi2014efficient}
Hajebi, K. and Zhang, H. (2014) An efficient index for visual search in
  appearance-based slam.
\newblock In \textit{IEEE Int. Conf. Robot. Autom.}, 353--358. IEEE.

\bibitem[{Han et~al.(2021)Han, Dong and Kan}]{han2021novel}
Han, J., Dong, R. and Kan, J. (2021) A novel loop closure detection method with
  the combination of points and lines based on information entropy.
\newblock \textit{Journal of Field Robotics}, \textbf{38}, 386--401.

\bibitem[{He et~al.(2016)He, Zhang, Ren and Sun}]{he2016deep}
He, K., Zhang, X., Ren, S. and Sun, J. (2016) Deep residual learning for image
  recognition.
\newblock 770--778.

\bibitem[{Hou et~al.(2015)Hou, Zhang and Zhou}]{hou2015convolutional}
Hou, Y., Zhang, H. and Zhou, S. (2015) Convolutional neural network-based image
  representation for visual loop closure detection.
\newblock In \textit{IEEE Int. Conf. Infor. Autom.}, 2238--2245.

\bibitem[{J{\'e}gou and Chum(2012)}]{jegou2012negative}
J{\'e}gou, H. and Chum, O. (2012) Negative evidences and co-occurences in image
  retrieval: The benefit of pca and whitening.
\newblock In \textit{Euro. Conf. Comp. Vis.}, 774--787.

\bibitem[{Jegou et~al.(2011)Jegou, Douze and Schmid}]{jegou2011product}
Jegou, H., Douze, M. and Schmid, C. (2011) Product quantization for nearest
  neighbor search.
\newblock \textit{IEEE Trans. Pattern Analysis and Machine Intell.},
  \textbf{33}, 117--128.

\bibitem[{J{\'e}gou et~al.(2010)J{\'e}gou, Douze, Schmid and
  P{\'e}rez}]{jegou2010aggregating}
J{\'e}gou, H., Douze, M., Schmid, C. and P{\'e}rez, P. (2010) Aggregating local
  descriptors into a compact image representation.
\newblock In \textit{IEEE Conf. Comp. Vis. Pattern Recogn.}, 3304--3311. IEEE.

\bibitem[{Kazmi and Mertsching(2019)}]{kazmi2019detecting}
Kazmi, S. A.~M. and Mertsching, B. (2019) Detecting the expectancy of a place
  using nearby context for appearance-based mapping.
\newblock \textit{IEEE Trans. Robot.}, \textbf{35}, 1352--1366.

\bibitem[{{Khan} and {Wollherr}(2015)}]{khan2015ibuild}
{Khan}, S. and {Wollherr}, D. (2015) {IBuILD: Incremental bag of binary words
  for appearance based loop closure detection}.
\newblock In \textit{IEEE Int. Conf. Robot. Autom.}, 5441--5447.

\bibitem[{Klein and Murray(2007)}]{klein2007parallel}
Klein, G. and Murray, D. (2007) Parallel tracking and mapping for small ar
  workspaces.
\newblock In \textit{IEEE and ACM Int. Symp. Mixed. Aug. Real.}, 225--234.

\bibitem[{Kleinberg(2000)}]{kleinberg2000navigation}
Kleinberg, J.~M. (2000) Navigation in a small world.
\newblock \textit{Nature}, \textbf{406}, 845.

\bibitem[{Konstantinidis et~al.(2005)Konstantinidis, Gasteratos and
  Andreadis}]{konstantinidis2005image}
Konstantinidis, K., Gasteratos, A. and Andreadis, I. (2005) Image retrieval
  based on fuzzy color histogram processing.
\newblock \textit{Opt. Commun.}, \textbf{248}, 375--386.

\bibitem[{Kostavelis and Gasteratos(2015)}]{kostavelis2015semantic}
Kostavelis, I. and Gasteratos, A. (2015) Semantic mapping for mobile robotics
  tasks: A survey.
\newblock \textit{Robotics and Autonomous Systems}, \textbf{66}, 86--103.

\bibitem[{Krizhevsky et~al.(2012)Krizhevsky, Sutskever and
  Hinton}]{krizhevsky2012imagenet}
Krizhevsky, A., Sutskever, I. and Hinton, G.~E. (2012) Imagenet classification
  with deep convolutional neural networks.
\newblock In \textit{Adv. Neural Inf. Process. Syst.}, 1097--1105.

\bibitem[{Labbe and Michaud(2013)}]{labbe2013appearance}
Labbe, M. and Michaud, F. (2013) Appearance-based loop closure detection for
  online large-scale and long-term operation.
\newblock \textit{IEEE Trans. Robot.}, \textbf{29}, 734--745.

\bibitem[{Lin et~al.(2013)Lin, Chen and Yan}]{lin2013network}
Lin, M., Chen, Q. and Yan, S. (2013) Network in network.
\newblock \textit{arXiv preprint arXiv:1312.4400}.

\bibitem[{Liu and Zhang(2012)}]{liu2012indexing}
Liu, Y. and Zhang, H. (2012) Indexing visual features: Real-time loop closure
  detection using a tree structure.
\newblock In \textit{IEEE Int. Conf. Robot. Autom.}, 3613--3618.

\bibitem[{Lowe(2004)}]{lowe2004distinctive}
Lowe, D.~G. (2004) Distinctive image features from scale-invariant keypoints.
\newblock \textit{Int. J. of Comp. Vis.}, \textbf{60}, 91--110.

\bibitem[{{Lowry} et~al.(2016){Lowry}, {SÃ¼nderhauf}, {Newman}, {Leonard},
  {Cox}, {Corke} and {Milford}}]{lowry2016visual}
{Lowry}, S., {SÃ¼nderhauf}, N., {Newman}, P., {Leonard}, J.~J., {Cox}, D.,
  {Corke}, P. and {Milford}, M.~J. (2016) Visual place recognition: A survey.
\newblock \textit{IEEE Trans. Robot.}, \textbf{32}, 1--19.

\bibitem[{MacQueen et~al.(1967)}]{macqueen1967some}
MacQueen, J. et~al. (1967) Some methods for classification and analysis of
  multivariate observations.
\newblock In \textit{roc. Berkeley Symp. Math. Statist. Prob.}, vol.~1,
  281--297.

\bibitem[{Malkov and Yashunin(2018)}]{malkov2018efficient}
Malkov, Y.~A. and Yashunin, D.~A. (2018) Efficient and robust approximate
  nearest neighbor search using hierarchical navigable small world graphs.
\newblock \textit{IEEE Trans. Pattern Analysis and Machine Intell.}

\bibitem[{Mei et~al.(2010)Mei, Sibley and Newman}]{mei2010closing}
Mei, C., Sibley, G. and Newman, P. (2010) Closing loops without places.
\newblock In \textit{IEEE/RSJ Int. Conf. Intell. Robots Syst.}, 3738--3744.

\bibitem[{Muja and Lowe(2009)}]{muja2009fast}
Muja, M. and Lowe, D.~G. (2009) Fast approximate nearest neighbors with
  automatic algorithm configuration.
\newblock \textit{Int. Conf. Comp. Vis. Theo. Appl.}, \textbf{2}, 2.

\bibitem[{Muja and Lowe(2014)}]{muja2014scalable}
--- (2014) Scalable nearest neighbor algorithms for high dimensional data.
\newblock \textit{IEEE Trans. Pattern Analysis and Machine Intell.},
  2227--2240.

\bibitem[{Mur-Artal and Tard{\'o}s(2014)}]{mur2014fast}
Mur-Artal, R. and Tard{\'o}s, J.~D. (2014) {Fast relocalisation and loop
  closing in keyframe-based SLAM}.
\newblock In \textit{IEEE Int. Conf. Robot. Autom.}, 846--853.

\bibitem[{{Nicosevici} and {Garcia}(2012)}]{nicosevici2012automatic}
{Nicosevici}, T. and {Garcia}, R. (2012) Automatic visual bag-of-words for
  online robot navigation and mapping.
\newblock \textit{IEEE Trans. Robot.}, \textbf{28}, 886--898.

\bibitem[{Noh et~al.(2017)Noh, Araujo, Sim, Weyand and Han}]{noh2017large}
Noh, H., Araujo, A., Sim, J., Weyand, T. and Han, B. (2017) Large-scale image
  retrieval with attentive deep local features.
\newblock In \textit{IEEE Conf. Comp. Vis.}, 3456--3465.

\bibitem[{Oliva and Torralba(2001)}]{oliva2001modeling}
Oliva, A. and Torralba, A. (2001) Modeling the shape of the scene: A holistic
  representation of the spatial envelope.
\newblock \textit{Int. J. of Comp. Vis.}, \textbf{42}, 145--175.

\bibitem[{Oliva and Torralba(2006)}]{oliva2006building}
--- (2006) {Building the gist of a scene: The role of global image features in
  recognition}.
\newblock \textit{Progr. Brain Res.}, \textbf{155}, 23--36.

\bibitem[{Radenovic et~al.(2018)Radenovic, Iscen, Tolias, Avrithis and
  Chum}]{radenovic2018revisiting}
Radenovic, F., Iscen, A., Tolias, G., Avrithis, Y. and Chum, O. (2018)
  Revisiting oxford and paris: Large-scale image retrieval benchmarking.
\newblock In \textit{IEEE Conf. Comp. Vis. Pattern Recogn.}, 5706--5715.

\bibitem[{Revaud et~al.(2019)Revaud, Almaz{\'a}n, Rezende and
  Souza}]{revaud2019learning}
Revaud, J., Almaz{\'a}n, J., Rezende, R.~S. and Souza, C. R.~d. (2019) Learning
  with average precision: Training image retrieval with a listwise loss.
\newblock In \textit{IEEE Conf. Comp. Vis.}, 5107--5116.

\bibitem[{Rublee et~al.(2011)Rublee, Rabaud, Konolige and
  Bradski}]{rublee2011orb}
Rublee, E., Rabaud, V., Konolige, K. and Bradski, G. (2011) Orb: An efficient
  alternative to sift or surf.
\newblock In \textit{IEEE Conf. Comp. Vis.}, 2564--2571.

\bibitem[{Russakovsky et~al.(2015)Russakovsky, Deng, Su, Krause, Satheesh, Ma,
  Huang, Karpathy, Khosla, Bernstein et~al.}]{russakovsky2015imagenet}
Russakovsky, O., Deng, J., Su, H., Krause, J., Satheesh, S., Ma, S., Huang, Z.,
  Karpathy, A., Khosla, A., Bernstein, M. et~al. (2015) Imagenet large scale
  visual recognition challenge.
\newblock \textit{Int. J. of Comp. Vis.}, \textbf{115}, 211--252.

\bibitem[{Sandler et~al.(2018)Sandler, Howard, Zhu, Zhmoginov and
  Chen}]{sandler2018mobilenetv2}
Sandler, M., Howard, A., Zhu, M., Zhmoginov, A. and Chen, L.-C. (2018)
  Mobilenetv2: Inverted residuals and linear bottlenecks.
\newblock In \textit{IEEE/IVF Conf. Comp. Vis. Pattern Recogn.}, 4510--4520.
  IEEE.

\bibitem[{Sivic and Zisserman(2003)}]{sivic2003video}
Sivic, J. and Zisserman, A. (2003) Video google: A text retrieval approach to
  object matching in videos.
\newblock In \textit{IEEE Conf. Comp. Vis.}, 1470.

\bibitem[{Smith et~al.(2009)Smith, Baldwin, Churchill, Paul and
  Newman}]{smith2009new}
Smith, M., Baldwin, I., Churchill, W., Paul, R. and Newman, P. (2009) The new
  college vision and laser data set.
\newblock \textit{Int. J. Robot. Res.}, \textbf{28}, 595--599.

\bibitem[{S{\"u}nderhauf et~al.(2015)S{\"u}nderhauf, Shirazi, Dayoub, Upcroft
  and Milford}]{sunderhauf2015performance}
S{\"u}nderhauf, N., Shirazi, S., Dayoub, F., Upcroft, B. and Milford, M. (2015)
  On the performance of convnet features for place recognition.
\newblock In \textit{IEEE/RSJ Int. Conf. Intell. Robots Syst.}, 4297--4304.

\bibitem[{Teichmann et~al.(2019)Teichmann, Araujo, Zhu and
  Sim}]{teichmann2019detect-to-retrieve}
Teichmann, M., Araujo, A., Zhu, M. and Sim, J. (2019) Detect-to-retrieve:
  Efficient regional aggregation for image search.
\newblock In \textit{IEEE Conf. Comp. Vis. Pattern Recogn.}, 5109--5118.

\bibitem[{Torr and Murray(1997)}]{torr1997development}
Torr, P.~H. and Murray, D.~W. (1997) The development and comparison of robust
  methods for estimating the fundamental matrix.
\newblock \textit{Int. J. Comput. Vis.}, \textbf{24}, 271--300.

\bibitem[{Torralba et~al.(2003)Torralba, Murphy, Freeman, Rubin
  et~al.}]{torralba2003context}
Torralba, A., Murphy, K.~P., Freeman, W.~T., Rubin, M.~A. et~al. (2003)
  Context-based vision system for place and object recognition.
\newblock In \textit{IEEE Conf. Comp. Vis.}, vol.~3, 273--280.

\bibitem[{{Tsintotas} et~al.(2018){Tsintotas}, {Bampis} and
  {Gasteratos}}]{tsintotas2018assigning}
{Tsintotas}, K.~A., {Bampis}, L. and {Gasteratos}, A. (2018) Assigning visual
  words to places for loop closure detection.
\newblock In \textit{IEEE Int. Conf. Robot. Autom.}, 5979--5985.

\bibitem[{Tsintotas et~al.(2018{\natexlab{a}})Tsintotas, Bampis and
  Gasteratos}]{tsintotas2018doseqslam}
Tsintotas, K.~A., Bampis, L. and Gasteratos, A. (2018{\natexlab{a}})
  {DOSeqSLAM: Dynamic on-line sequence based loop closure detection algorithm
  for SLAM}.
\newblock In \textit{IEEE Int. Conf. Imag. Sys. Techn.}, 1--6.

\bibitem[{{Tsintotas} et~al.(2019){Tsintotas}, {Bampis} and
  {Gasteratos}}]{tsintotasRAL}
{Tsintotas}, K.~A., {Bampis}, L. and {Gasteratos}, A. (2019) Probabilistic
  appearance-based place recognition through bag of tracked words.
\newblock \textit{IEEE Robot. Autom. Lett.}, \textbf{4}, 1737--1744.

\bibitem[{Tsintotas et~al.(2021)Tsintotas, Bampis and
  Gasteratos}]{tsintotas2021modest}
Tsintotas, K.~A., Bampis, L. and Gasteratos, A. (2021) Modest-vocabulary
  loop-closure detection with incremental bag of tracked words.
\newblock \textit{Robotics and Autonomous Systems}, \textbf{141}, 103782.

\bibitem[{Tsintotas et~al.(2018{\natexlab{b}})Tsintotas, Bampis, Rallis and
  Gasteratos}]{tsintotas2018seqslam}
Tsintotas, K.~A., Bampis, L., Rallis, S. and Gasteratos, A.
  (2018{\natexlab{b}}) {SeqSLAM with bag of visual words for appearance based
  loop closure detection}.
\newblock In \textit{Proc. Int. Conf. Robot. in Alpe-Adria Danube Region.},
  580--587.

\bibitem[{Tsintotas et~al.(2019)Tsintotas, Giannis, Bampis and
  Gasteratos}]{tsintotas2019appearance}
Tsintotas, K.~A., Giannis, P., Bampis, L. and Gasteratos, A. (2019)
  Appearance-based loop closure detection with scale-restrictive visual
  features.
\newblock In \textit{Proc. Int. Conf. Comput. Vis. Syst.}, 75--87.

\bibitem[{Wang et~al.(2018)Wang, Huang, Lin, Hu, Zeng and
  Sun}]{wang2018omnidirectional}
Wang, T.-H., Huang, H.-J., Lin, J.-T., Hu, C.-W., Zeng, K.-H. and Sun, M.
  (2018) Omnidirectional cnn for visual place recognition and navigation.
\newblock In \textit{IEEE Int. Conf. Robot. Autom.}, 2341--2348.

\bibitem[{Weyand et~al.(2020)Weyand, Araujo, Cao and Sim}]{weyand2020google}
Weyand, T., Araujo, A., Cao, B. and Sim, J. (2020) Google landmarks dataset
  v2--a large-scale benchmark for instance-level recognition and retrieval.
\newblock \textit{arXiv preprint arXiv:2004.01804}.

\bibitem[{Xia et~al.(2016)Xia, Li, Qi and Fan}]{xia2016loop}
Xia, Y., Li, J., Qi, L. and Fan, H. (2016) Loop closure detection for visual
  slam using pcanet features.
\newblock In \textit{IEEE Joint Conf. Neur. Net.}, 2274--2281.

\bibitem[{Xie et~al.(2017)Xie, Girshick, Doll{\'a}r, Tu and
  He}]{xie2017aggregated}
Xie, S., Girshick, R., Doll{\'a}r, P., Tu, Z. and He, K. (2017) Aggregated
  residual transformations for deep neural networks.
\newblock In \textit{IEEE Conf. Comp. Vis. Pattern Recogn.}, 1492--1500.

\bibitem[{Xin et~al.(2019)Xin, Cui, Zhang, Yang and Wang}]{xin2019real}
Xin, Z., Cui, X., Zhang, J., Yang, Y. and Wang, Y. (2019) Real-time visual
  place recognition based on analyzing distribution of multi-scale cnn
  landmarks.
\newblock \textit{Journal of Intelligent \& Robotic Systems}, \textbf{94},
  777--792.

\bibitem[{Yu et~al.(2020)Yu, Zhu, Zhang, Huang and Tao}]{yu2019spatial}
Yu, J., Zhu, C., Zhang, J., Huang, Q. and Tao, D. (2020) Spatial
  pyramid-enhanced netvlad with weighted triplet loss for place recognition.
\newblock \textit{IEEE Trans. Neur. Net. and Learn. Syst.}, \textbf{31},
  661--674.

\bibitem[{Zhang(2011)}]{zhang2011borf}
Zhang, H. (2011) {BoRF: Loop-closure detection with scale invariant visual
  features}.
\newblock In \textit{IEEE Int. Conf. Robot. Autom.}, 3125--3130.

\bibitem[{Zhang et~al.(2020)Zhang, Wu, Zhang, Zhu, Zhang, Lin, Sun, He,
  Mueller, Manmatha et~al.}]{zhang2020resnest}
Zhang, H., Wu, C., Zhang, Z., Zhu, Y., Zhang, Z., Lin, H., Sun, Y., He, T.,
  Mueller, J., Manmatha, R. et~al. (2020) Resnest: Split-attention networks.
\newblock \textit{arXiv preprint arXiv:2004.08955}.

\bibitem[{Zhou et~al.(2014)Zhou, Lapedriza, Xiao, Torralba and
  Oliva}]{zhou2014learning}
Zhou, B., Lapedriza, A., Xiao, J., Torralba, A. and Oliva, A. (2014) Learning
  deep features for scene recognition using places database.
\newblock In \textit{Adv. Neural Inf. Process. Syst.}, 487--495.

\end{thebibliography}



\end{document}